\documentclass[a4paper,fleqn]{cas-sc}
\makeatletter
\def\printorcid{\relax}
\makeatother

\usepackage{float}
\usepackage{siunitx}

\usepackage{subcaption} % in your preamble
\usepackage[numbers]{natbib}
\usepackage{amsthm}
\usepackage[ruled,vlined,linesnumbered]{algorithm2e}
\SetKwInOut{Input}{Input}
\SetKwInOut{Output}{Output}
\theoremstyle{plain}
\newtheorem{theorem}{Theorem}
\newtheorem{lemma}{Lemma}

\theoremstyle{definition}
\newtheorem{definition}[theorem]{Definition}
\newtheorem{assumption}{Assumption}
\theoremstyle{remark}

\def\tsc#1{\csdef{#1}{\textsc{\lowercase{#1}}\xspace}}
\tsc{WGM}
\tsc{QE}
\begin{document}
\let\WriteBookmarks\relax
\def\floatpagepagefraction{1}
\def\textpagefraction{.001}

\shorttitle{GC-Fed: Gradient Centralized Federated Learning}    

\shortauthors{}  

\title [mode = title]{ GC-Fed: Gradient Centralized Federated Learning with Partial Client Participation}  


% \tnotetext[1]{Gradient Centralized Federated Learning} 


\author[1]{Jungwon Seo}%[<options>]

\ead{jungwon.seo@uis.no}


\credit{Writing – original draft, Methodology, 
Writing – review \& editing}
\affiliation[1]{
            organization={Department of Electrical Engineering and Computer Science, University of Stavanger},
            city={Stavanger},
%          citysep={}, % Uncomment if no comma needed between city and postcode
            postcode={4021}, 
            country={Norway}}

% \fntext[1]{First author}
\author[1]{Ferhat Ozgur Catak}%[<options>]
\ead{f.ozgur.catak@uis.no}
\credit{Writing – review \& editing}
\author[1]{Chunming Rong}%[<options>]
\ead{chunming.rong@uis.no}
\credit{Writing – review \& editing, Funding acquisition}

\author[2]{Kibeom Hong}%[<options>]
\cormark[1]
% Footnote of the second author

% Email id of the second author
\ead{kb.hong@sookmyung.ac.kr}


\credit{Writing – original draft,
Writing – review \& editing}

\affiliation[2]{organization={Department of Software, Sookmyung Women's University},
            city={Seoul},
%          citysep={}, % Uncomment if no comma needed between city and postcode
            postcode={04310}, 
            country={South Korea}}

\cortext[1]{Corresponding authors}
\author[3]{Minhoe Kim}%[<options>]
\ead{kimminhoe@seoultech.ac.kr}
\cormark[1]
\affiliation[3]{organization={Department of Electrical and Information Engineering, Seoul National University of Science and Technology},
            city={Seoul},
%          citysep={}, % Uncomment if no comma needed between city and postcode
            postcode={01811}, 
            country={South Korea}}
\credit{Writing – original draft,
Writing – review \& editing, Supervision}

\begin{abstract}
Federated Learning (FL) enables privacy-preserving multi-source information fusion (MSIF) but is challenged by client drift in highly heterogeneous data settings. Many existing drift-mitigation strategies rely on reference-based techniques—such as gradient adjustments or proximal loss—that use historical snapshots (e.g., past gradients or previous global models) as reference points. When only a subset of clients participates in each training round, these historical references may not accurately capture the overall data distribution, leading to unstable training. In contrast, our proposed Gradient Centralized Federated Learning (GC-Fed) employs a hyperplane as a historically independent reference point to guide local training and enhance inter-client alignment. GC-Fed comprises two complementary components: Local GC, which centralizes gradients during local training, and Global GC, which centralizes updates during server aggregation. In our hybrid design, Local GC is applied to feature-extraction layers to harmonize client contributions, while Global GC refines classifier layers to stabilize round-wise performance. Theoretical analysis and extensive experiments on benchmark FL tasks demonstrate that GC-Fed effectively mitigates client drift and achieves up to a 20\% improvement in accuracy under heterogeneous and partial participation conditions.

\end{abstract}



% Keywords

\maketitle

% Main text
\section{Introduction}\label{}

Federated Learning (FL) embodies a privacy-preserving paradigm that harnesses multi-source information fusion (MSIF)~\cite{zhang2021multi,ji2024emerging}  to enable distributed training~\cite{mcmahan2017communication}. In this framework, each client independently extracts distinctive insights from its local dataset, converting these observations into model updates—whether in the form of complete models or incremental parameter changes. These locally derived updates, representing diverse and heterogeneous data sources, are then transmitted to a central server. Here, a fusion process, often relying on simple averaging~\cite{mcmahan2017communication} or more sophisticated consensus strategies~\cite{li2020federated,karimireddy2020scaffold,jhunjhunwala2022fedvarp,lee2024fedsol,lee2022preservation}, integrates these contributions into a unified global model. This collaborative synthesis not only upholds data privacy but also capitalizes on the richness of multi-source information~\cite{xinde2024multi,huang2024multimodal}, resulting in a more robust and comprehensive representation of the underlying data landscape and ultimately driving enhanced model refinement.

Despite the promise of FL for harnessing multi-source data while preserving data privacy, several challenges persist. A primary concern is the intrinsic heterogeneity of client data. In practice, there is no guarantee that data held by individual clients is homogeneous; indeed, most multi-source datasets are inherently diverse. This heterogeneity is evident not only in multi-modal scenarios but also in single-modality settings, where differences in data volume, feature representation, and label distribution are common. Under the assumption of independent and identically distributed (\textit{i.i.d.}) client data, FL can achieve performance comparable to traditional distributed learning approaches \cite{chu2006map,zinkevich2010parallelized,dean2012large}. However, when client data are \textit{non-i.i.d.}, such disparities lead to a phenomenon known as client drift~\cite{karimireddy2020scaffold}. In this context, each client traverses a distinct loss landscape during optimization, resulting in divergent local training trajectories and convergence toward different optima \cite{al-shedivat2021federated,zhang2022federated,seo2024understanding}. These issues not only slow down the optimization process but also undermine the stability and robustness of the global model.

Addressing client drift in FL has spurred various approaches aimed at aligning update directions and model weights across clients. A widely adopted strategy incorporates an auxiliary proximal term into the task-specific loss~\cite{li2020federated,acar2021federated,li2021model,lee2022preservation,lee2024fedsol}. This term is designed to account for discrepancies among client updates, facilitating the simultaneous minimization of both the primary loss (e.g., cross-entropy) and various factors influencing client drift. An alternative line of work focuses on directly correcting gradients or local updates, either during client-side training or at the global aggregation step~\cite{karimireddy2020scaffold,haddadpour2021federated,acar2021federated,kim2024communication}. These methods, often inspired by classical optimization techniques, are adapted to the FL setting to mitigate both client-wise and round-wise variance.

\begin{figure}[htbp]
  \centering
  \subfloat[Visualization of Partial Participation Effects]{%
    \includegraphics[width=0.43\linewidth]{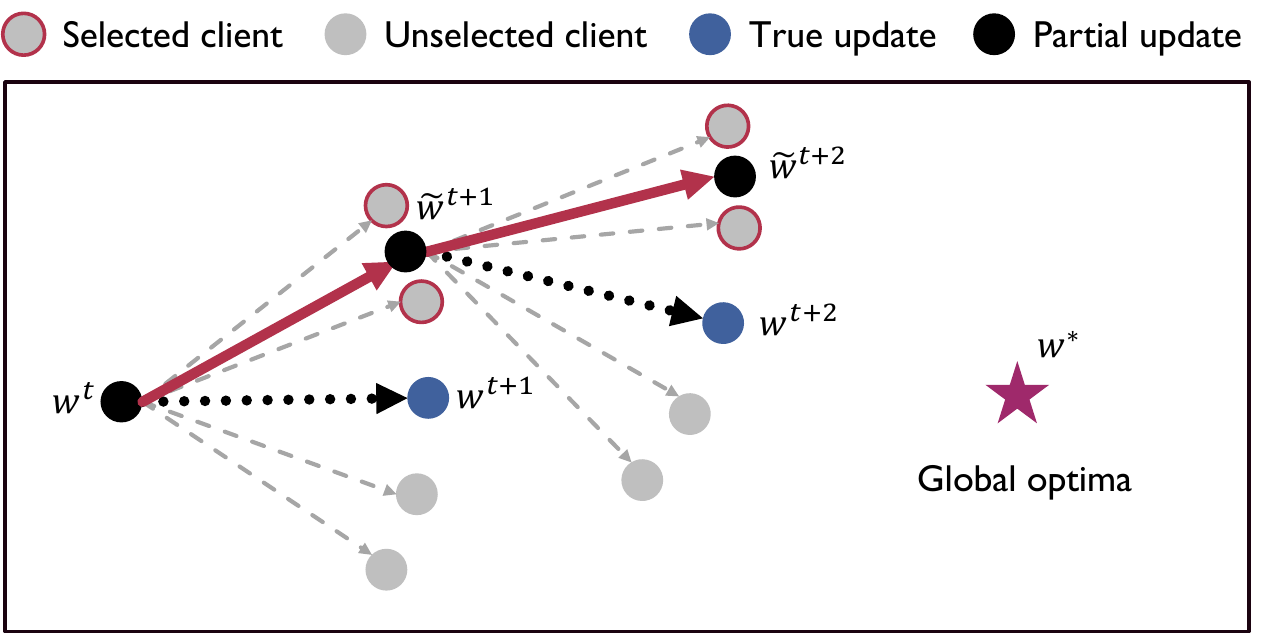}%
    \label{fig:partial-participation-a}
  }
  \hfill
  \subfloat[Accuracy and Update Discrepancy Across Varying Client Participation]{%
    \includegraphics[width=0.52\linewidth]{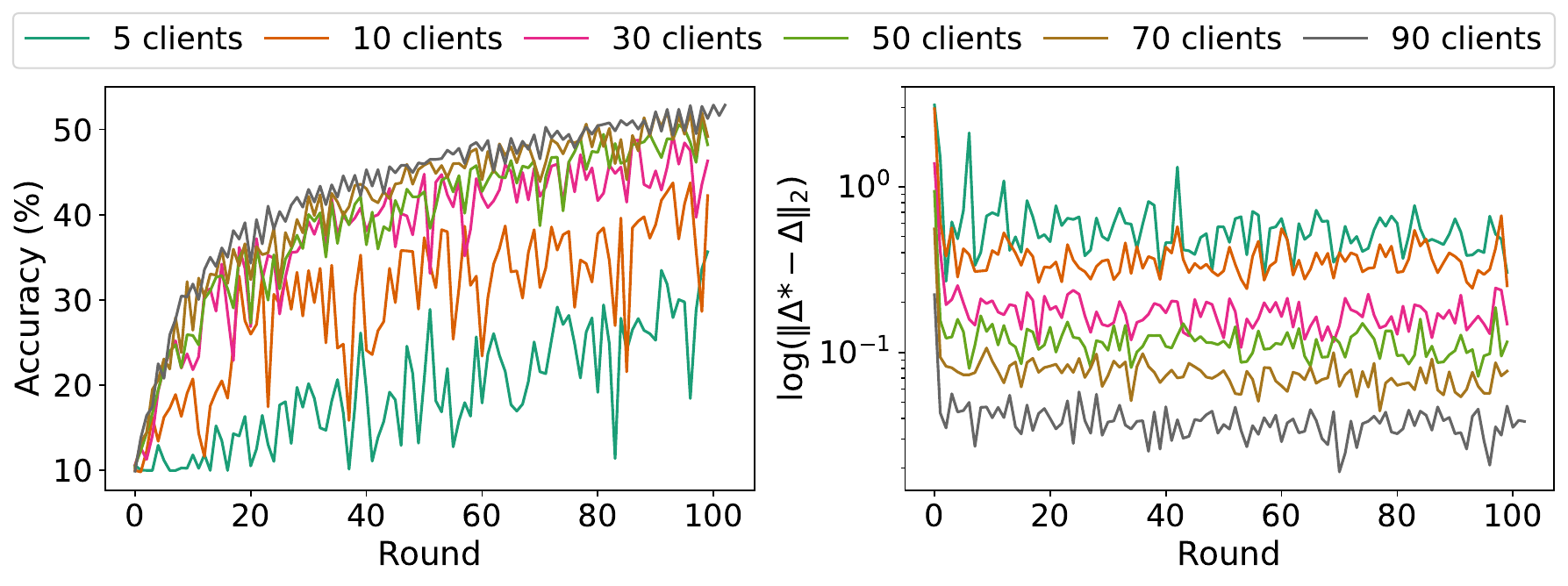}%
    \label{fig:partial-participation-b}
  }
  \caption{In partial participation scenarios, the gap between the true update (aggregated from all clients) and the partial update (aggregated from a subset of clients) widens as the number of participating clients decreases, leading to increased training instability, unlike in an \textit{i.i.d.} setting.}
  \label{fig:partial-participation}
\end{figure}

Prior approaches to adjusting local training typically leverage historical information—such as the previous round’s global model, client updates, or control variates—to guide the next round of optimization.
These methods have proven effective in cross-silo settings (e.g., data centers), where a majority of clients participate consistently. However, in cross-device scenarios (e.g., IoT devices and smartphones), where only a small fraction of thousands or millions of clients participate per round due to client availability and server hardware limitations, the sampled subset may fail to accurately reflect the optimal update direction~\cite{jhunjhunwala2022fedvarp,seo2024relaxed}. As illustrated in Figure~\ref{fig:partial-participation-a}, partial client participation can yield a global update that deviates from the one obtained via full participation. Moreover, our preliminary experiments (Figure~\ref{fig:partial-participation-b}) demonstrate that the discrepancy between the true update (aggregated from all clients) and the partial update increases as participation diminishes. In other words, the historical information generated in previous rounds—even when not limited to the global model—can be less effective in guiding the current update when the participating clients do not reliably represent the entire population.

\begin{figure}[htbp]
    \centering
    \includegraphics[width=0.6\linewidth]{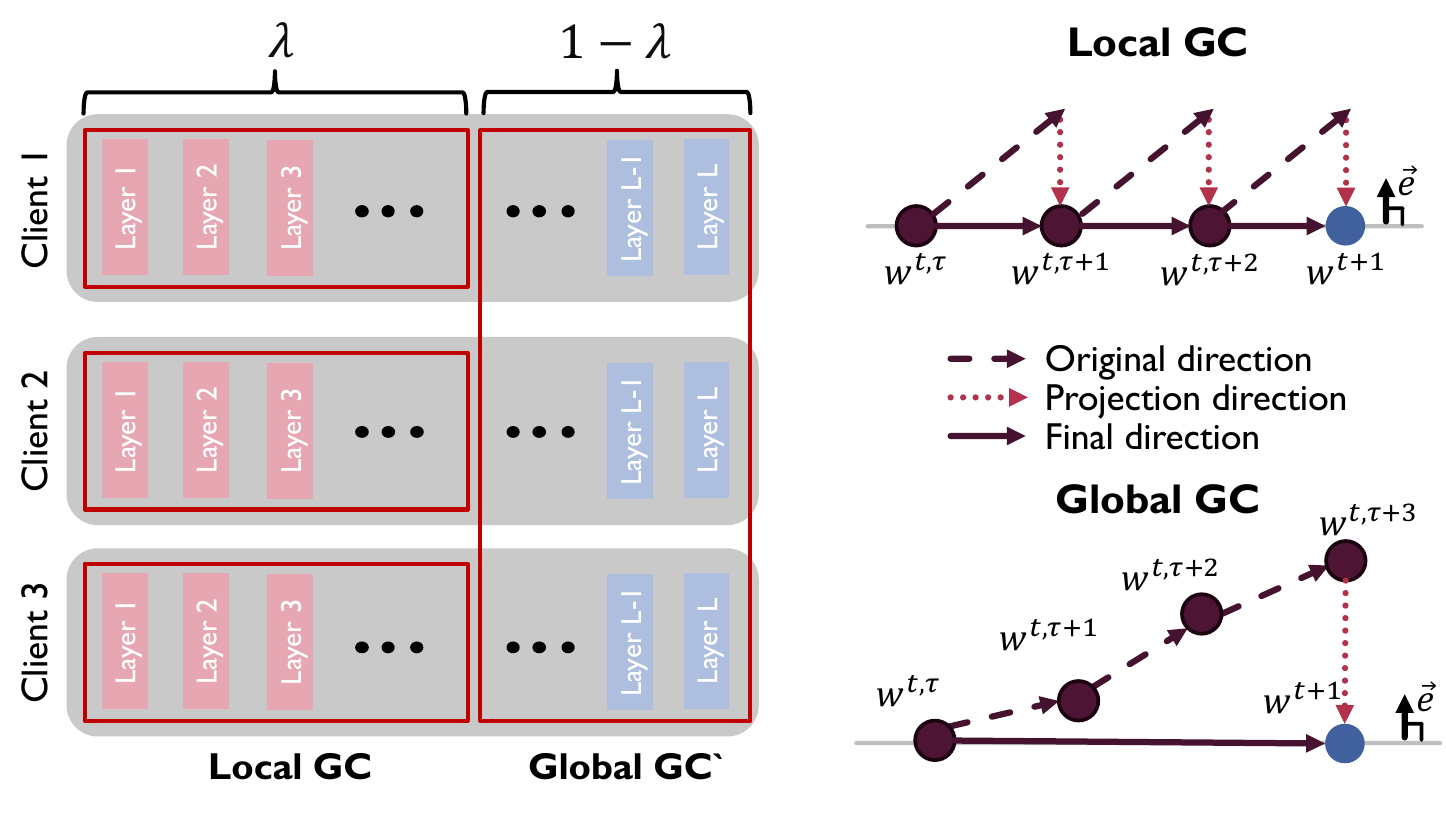}
    \caption{Overview of the proposed \texttt{GC-Fed} framework for FL. Our approach incorporates Gradient Centralization in two phases: \texttt{Local GC} is applied to the feature extraction layers during local SGD, while \texttt{Global GC} is employed at the classifier layer during model aggregation. The application of these strategies is modulated by the hyperparameter $\lambda$.}
    \label{fig:gc-fed-architecture}
\end{figure}

To address this challenge, we propose a method that promotes the alignment of local client updates while preserving their core update direction. Specifically, we introduce a historically independent yet shared reference via Gradient Centralization (GC)~\cite{yong2020gradient}. GC was originally designed to enforce a zero-mean distribution on the out-channel gradients of layer-specific matrices to improve training stability and performance. GC can also be interpreted as a projection-based technique that maps gradients onto a fixed hyperplane~\cite{yong2020gradient}. As a result, we contend that this hyperplane can serve as a stable reference for update correction across both clients and rounds in FL. 

Motivated by this insight, we first examine two straightforward variants: (i) \texttt{Local GC}, performed during local optimization, and (ii) \texttt{Global GC}, conducted as part of the server-side aggregation of local updates, similar to other server-side optimization algorithms.~\cite{hsu2019measuring,reddi2021adaptive,jhunjhunwala2022fedvarp,kim2024communication}. Preliminary experiments reveal that while both variants outperform \texttt{FedAvg}, \texttt{Local GC} attains a higher peak performance at the cost of significant round-to-round fluctuations, and \texttt{Global GC} yields more stable updates with a slightly lower peak performance. Recognizing the trade-offs between performance and stability, we introduce a hybrid approach called \texttt{GC-Fed}. Specifically, we leverage the layer-wise roles in deep neural networks by applying \texttt{Local GC} to the feature extraction layers and \texttt{Global GC} to the classifier layers (see Figure~\ref{fig:gc-fed-architecture}). This design maintains the high performance of \texttt{Local GC} while leveraging the stability provided by \texttt{Global GC}. Moreover, by centralizing only the last layer during the global stage, the inter-layer alignment approaches are observed under the nearly \textit{i.i.d.} setting of \texttt{FedAvg}. Extensive experiments demonstrate that \texttt{GC-Fed} outperforms current state-of-the-art methods, achieving robust and consistent improvements even under partial client participation. Our official source code can be found in \href{https://github.com/thejungwon/GC-Fed}{https://github.com/thejungwon/GC-Fed} for reproducibility.

Our contributions are summarized as follows: 
\begin{itemize} 
\item We demonstrate that gradient centralization can effectively reduce client variance in FL by providing a shared projection plane. 
\item We propose \texttt{GC-Fed}, a novel FL algorithm that flexibly integrates GC into both the local and global optimization stages. 
\item We validate \texttt{GC-Fed} through extensive experiments, showing its superiority over conventional algorithms across various datasets and settings, supported by both theoretical and empirical analysis. 
\end{itemize}

\section{Related Works}
\label{sec:relatedwork}

\subsection{Federated Learning Overview} Since the introduction of \texttt{FedAvg}~\cite{mcmahan2017communication} as the foundational algorithm for FL, numerous studies have pointed out its performance degradation in \textit{non-i.i.d.} environments and proposed a variety of approaches driven by diverse motivations. Early research identified divergences in local client optimization, primarily due to differences in data distribution and volume across clients. To address this, \texttt{FedProx}~\cite{li2020federated} introduces a proximal term in the local loss function, constraining local model updates to remain closer to the global model. \texttt{SCAFFOLD}~\cite{karimireddy2020scaffold} defines this divergence as client drift and corrects local gradients using a control variate. More recent approaches focus on server-side techniques to mitigate client drift. \texttt{FedOpt}~\cite{reddi2021adaptive} incorporates optimizers like Adam~\cite{kingma2014adam}, Yogi~\cite{zaheer2018adaptive}, and Adagrad~\cite{duchi2011adaptive} into the aggregation process, allowing the server-side aggregation to function similarly to local optimizers, thus enhancing training performance. Several alternative approaches have been proposed to enhance FL from distinct perspectives. For example, \texttt{FedSAM}~\cite{qu2022generalized} improves generalization by incorporating Sharpness-Aware Minimization (SAM). This strategy encourages clients to explore flatter minima, which in turn is expected to reduce divergence among client models on the loss landscape. \texttt{FedNTD}~\cite{lee2022preservation} and \texttt{FedSOL}~\cite{lee2024fedsol} treat FL as a continual learning problem, addressing catastrophic forgetting by incorporating KL-divergence proximal loss between local and global models in the local loss function, or by applying the SAM optimizer with orthogonal learning perspective of weight perturbation.

\subsection{Variance Reduction Methods in FL}
In contemporary optimization, the shift from full-batch gradient descent to SGD and mini-batch SGD introduces gradient variance due to the stochastic nature of these methods, often hindering optimizer convergence~\cite{gower2020variance}. To counteract this, a range of strategies—such as momentum, adaptive learning rate approaches~\cite{qian1999momentum,duchi2011adaptive,kingma2014adam}, and stochastic variance reduction techniques like SAG, SVRG, and SAGA~\cite{schmidt2017minimizing,johnson2013accelerating,defazio2014saga,hofmann2015variance}—have been developed. 

In FL, variance reduction presents additional challenges beyond those in centralized optimization. Here, variance arises not only from individual data samples in local training but also from the aggregation of models trained across diverse clients~\cite{das2022faster,li2023effectiveness}. Consequently, reducing local model variance does not necessarily yield a corresponding reduction in variance during global aggregation. To address these challenges, FL-specific variance reduction techniques target inter-client variance. For example, \texttt{SCAFFOLD} extends SVRG to FL by using two control variates: one derived from each client’s local updates and another from the globally aggregated model. While several algorithms adopt this approach~\cite{li2020federated,karimireddy2020scaffold,acar2021federated}, relying on reference points from previous rounds can introduce inconsistency due to partial participation under \textit{non-i.i.d.} Furthermore, from a practical standpoint, maintaining these reference points increases communication costs~\cite{karimireddy2020scaffold,haddadpour2021federated}, requires clients to store past state information~\cite{tian2022fedfor}, or demands extra server memory for client-specific data~\cite{jhunjhunwala2022fedvarp}. This dependency makes such approaches less efficient than \texttt{FedAvg} in terms of communication and storage, especially in cross-device FL with large client numbers. To address this practical challenge, we propose a reference-free approach leveraging GC.

\subsection{Server-Side Optimization in FL}
Although many FL algorithms focus on improving performance on the client side, research on improving the server-side is also being proposed continuously. Moreover, server-side optimization techniques are useful from a development perspective because additional changes can be made only on the server without modifying the existing client code~\cite{lo2022architectural,seo2023flexible}. \texttt{FedAvgM}~\cite{hsu2019measuring} incorporates a momentum strategy within the aggregation process, and \texttt{FedOpt}~\cite{reddi2021adaptive} introduces a suite of techniques aimed at enhancing server-side optimization. Another notable approach, \texttt{FedVARP}~\cite{jhunjhunwala2022fedvarp}, employs a variant of SAGA~\cite{defazio2014saga} by retaining a memory of previous updates from all clients on the server, thereby mitigating the negative impacts of partial client participation in FL. More recently, \texttt{FedACG}~\cite{kim2024communication} introduced a server-side look-ahead mechanism designed to accelerate convergence without incurring additional communication overhead. 

These efforts collectively highlight that server-side optimization can significantly improve FL performance. In our proposed approach, we interpret local updates (i.e., accumulated gradients) as gradients from the server’s perspective, centralizing them during the aggregation phase for the classifier layer. Similar to how GC operates in local training, this centralization aligns local updates from the perspective of the global model, ultimately facilitating more robust and efficient convergence.

\subsection{Classfier Variance in FL}
In FL with deep neural networks, the classifier layer has emerged as a significant source of model variance across clients. Prior work~\cite{oh2021fedbabu, luo2021no, legate2023guiding, li2023effectiveness, fani2024accelerating} has primarily focused on aligning the classifier layer or its outputs, showing that refining this component alone can significantly enhance overall model performance in FL. This variance in the classifier layer arises from its unique role in mapping extracted features to specific classes, contrasting with the earlier layers that primarily capture general feature representations~\cite{li2023effectiveness,kalla2024robust}. The divergence in the classifier layer is especially pronounced in FL due to class imbalance of \textit{non-i.i.d.} data, where each client may lack certain classes or exhibit skewed distributions of class occurrences~\cite{zhang2022federated,he2024fedcpd}. To address this issue, we propose applying GC exclusively to the classifier layer during global aggregation while retaining GC in the earlier layers during local updates. This approach aims to improve inter-client classifier alignment effectively.

\section{Proposed Method}

\begin{table}[htbp]
\caption{Summary of Notations}
\centering
\resizebox{0.5\linewidth}{!}{
\small
\begin{tabular}{|cl|}
\hline
\textbf{Notation} & \textbf{Description} \\ \hline
$\mathbf{w}$ & Model parameters \\ 
$\Delta$ & Model update after local training \\ 
$\mathbf{G}$ & Local gradient \\ 
$\mathbf{P}$ & Gradient centralization operation (projection-based) \\ 
$\ell(\mathbf{w}; \xi_{k,j})$ & Loss function at $\mathbf{w}$ for data point $\xi_{k,j}$ \\ 
$\xi_{k,j}$ & $j$-th data point of the $k$-th client \\ 
$F(\mathbf{w})$ & Global objective function to minimize \\ 
$F_k(\mathbf{w})$ & Local objective for the $k$-th client \\
$p_k$ & Weight (contribution) for the $k$-th local objective \\
$n_k$ & Number of data points in the $k$-th client \\ 
$N$ & Total number of clients \\ 
$C$ & Client participation ratio \\ 
$K$ & Number of selected clients \\ 
$\mathcal{K}$ & Set of selected clients \\ 
$E$ & Number of local epochs \\ 
$R$ & Number of communication rounds \\ 
$\eta$ & Local SGD learning rate \\ 
$\alpha$ & Non-IID factor (based on LDA data partitioning) \\ 
$\lambda$ & Threshold layer for local vs. global centralization \\ 
\hline
\end{tabular}
}

\end{table}

\subsection{Preliminary: Gradient Centralization}\label{sec:gradient-centralization}

GC is an optimization technique designed to complement various normalization methods such as Batch Normalization, Group Normalization, and Layer Normalization~\cite{ioffe2015batch,ba2016layer,wu2018group}. Unlike these normalization techniques, which directly modify activation matrices, GC targets gradient matrices during backpropagation. By centralizing gradients, GC has been proven to enhance model performance and training stability through improved generalization and a smoother loss landscape~\cite{yong2020gradient}. 

We first review GC through its two equivalent formulations, both of which remove the mean from the gradients~\cite{yong2020gradient}. Let 
$\mathbf{G} \in \mathbb{R}^{m \times n}$ be the gradient matrix for a given layer's parameters, where \(m\) denotes the number of input channels (or kernel components, if applicable), and \(n\) denotes the number of output channels. Each element \(g_{ij}\) in \(\mathbf{G}\) represents the gradient of the loss with respect to the weight connecting the \(i\)-th input to the \(j\)-th output.

\noindent \paragraph{\textbf{Mean-Subtraction Formulation.}}

For each output channel \(j\), compute the mean gradient:
\begin{equation}\label{eq:mu_g}
\mu_{\mathbf{G},j} = \frac{1}{m}\sum_{i=1}^{m} g_{ij}.    
\end{equation}
Assemble these into the mean vector:
\[
\mu_{\mathbf{G}} = \begin{bmatrix} \mu_{\mathbf{G},1} & \mu_{\mathbf{G},2} & \cdots & \mu_{\mathbf{G},n} \end{bmatrix}.
\]
Then, subtract the mean from every row to obtain the centralized gradient:

\begin{equation}\label{eq:mean-subtract}
\tilde{\mathbf{G}} = \mathbf{G} - \mathbf{1}_m \mu_{\mathbf{G}}.
\end{equation}

\noindent \paragraph{\textbf{Projection Formulation.}}

Equation~\eqref{eq:mean-subtract} can be compactly rewritten using \(\mathbf{e} = \frac{1}{\sqrt{m}}\mathbf{1}_m\) as:
\[
\tilde{\mathbf{G}} = \mathbf{G} - \mathbf{e}\mathbf{e}^\top \mathbf{G}.
\]

This formulation reveals that the mean-subtraction step can alternatively be interpreted as a projection operation. Specifically, Equation~\eqref{eq:mean-subtract} shows that gradient centering is equivalent to projecting \(\mathbf{G}\) onto the hyperplane orthogonal to \(\mathbf{e}\). Define the projection matrix:
\begin{equation}\label{eq:projection-matrix}
\mathbf{P} = \mathbf{I}_m - \mathbf{e}\mathbf{e}^\top,
\end{equation}
so that the centralized gradient becomes:

\[
\tilde{\mathbf{G}} = \mathbf{P}\mathbf{G}.    
\]

Expressing GC via the projection matrix (Equation~\eqref{eq:projection-matrix}) clarifies that each client's gradients are projected onto a common hyperplane. This hyperplane can serve as a reliable reference for aligning local gradients or updates, potentially offering benefits in FL—even without additional information sharing. To simplify notation, we use the term projection expression for this centralization operation throughout the rest of the paper.

\subsection{Problem Formulation}
In the context of FL, we aim to minimize a global objective \( F(\mathbf{w}) \), which represents the weighted sum of local objective functions \( F_k(\mathbf{w}) \) across multiple clients. Each client \( k \) holds its own local dataset and computes its local objective function. The problem is formulated as follows:

\[
\min_{\mathbf{w}} F(\mathbf{w}) \triangleq \sum_{k=1}^N p_k F_k(\mathbf{w}),
\]

where \( \mathbf{w} \) denotes the global model parameters that are shared across clients, and \( p_k \) is the weight (often proportional to the size of each client’s dataset) associated with client \( k \), satisfying \( \sum_{k=1}^N p_k = 1 \). Here, \( N \) represents the total number of clients.

Each client \( k \) optimizes its local objective function \( F_k(\mathbf{w}) \) based on its own data:

\[
F_k(\mathbf{w}) \triangleq \frac{1}{n_k} \sum_{j=1}^{n_k} \ell(\mathbf{w}; \xi_{k,j}),
\]

where \( n_k \) denotes the number of data points held by client \( k \), and \( \ell(\mathbf{w}; \xi_{k,j}) \) is the loss function computed for model parameters \( \mathbf{w} \) on the \( j \)-th data point \( \xi_{k,j} \) in client \( k \)’s local dataset. The global objective \( F(\mathbf{w}) \) is thus a weighted average of the local objectives across all clients.

\subsection{Proposed Method: GC-Fed}
\begin{figure}[htbp] 
    \centering
    \includegraphics[width=0.5\linewidth]{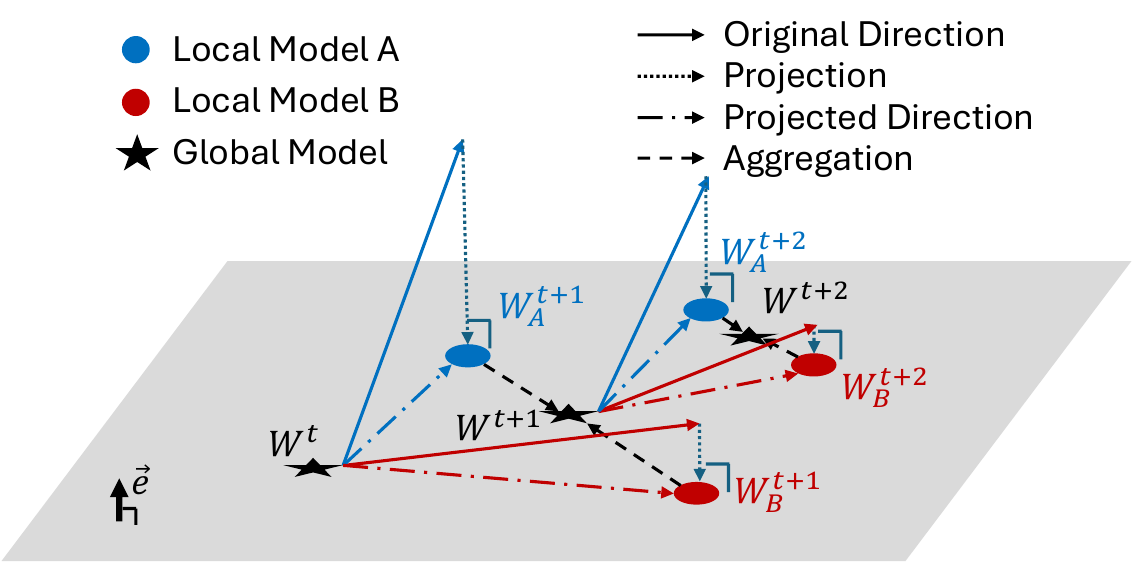}
\caption{Visualization of GC as a gradient projection method in FL}

\label{fig:gradient-projected-FL}
\end{figure}

The effect of GC in FL is visualized in Figure~\ref{fig:gradient-projected-FL}. Clients A and B initiate from the global model $\mathbf{w}^t$. For simplicity, a one-step gradient descent is assumed, where each client’s local model update is directed along a projected gradient path. The server then aggregates these updates to produce a new global model.

In this work, we first present two strategies for integrating GC into FL, followed by the introduction of our proposed method, \texttt{GC-Fed}. The first approach applies GC during local training, mirroring its usage in centralized learning.

\begin{definition}[\texttt{Local GC}]
Let \(\mathbf{G}\) denote the gradient computed locally by a client. The centralized local gradient, \(\Tilde{\mathbf{G}}\), is defined as
\[
\Tilde{\mathbf{G}} = \mathbf{P}\,\mathbf{G},
\]
\end{definition}
where \(\mathbf{P}\) is the projection matrix defined in Equation~\eqref{eq:projection-matrix}.

The second strategy explores the application of GC on the server-side, akin to global optimization techniques. Here, GC is applied to the aggregated local updates.

\begin{definition}[\texttt{Global GC}]
Let \(\Delta \) denote the aggregated update from selected clients. The centralized global update, \(\Tilde{\Delta}\), is defined as 
\[
\Tilde{\Delta } = \mathbf{P}\,\Delta ,
\]
where \(\mathbf{P}\) (from Equation~\eqref{eq:projection-matrix}) acts as a projection matrix on the averaged update.
\end{definition}

\begin{figure}[htbp]
    \centering
    \includegraphics[width=\linewidth]{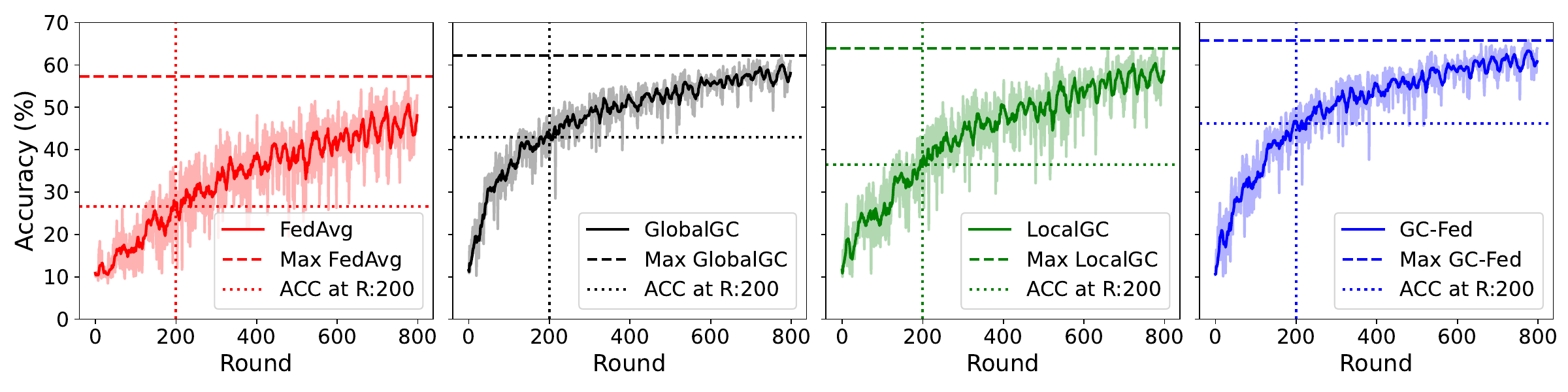}
\caption{Training dynamics of Top-1 test accuracy across different GC methods on CIFAR-10 with a CNN model (R: 800, participation: 5/200, LDA $\alpha = 0.05$). The curves are smoothed using a moving average window of 10, with shaded areas representing the original values. Dashed lines indicate the peak accuracy, while dotted vertical and horizontal lines mark the smoothed accuracy at round 200.}

    \label{fig:local-global-gc-performance}
\end{figure}

Empirical evaluations demonstrate that both \texttt{Local GC} and \texttt{Global GC} outperform \texttt{FedAvg}, though they exhibit distinct training dynamics. Figure~\ref{fig:local-global-gc-performance} illustrates test accuracy trends, where dashed lines indicate peak accuracy and cross-dots represent the smoothed accuracy at round 200. Notably, while both methods surpass \texttt{FedAvg}, \texttt{FedAvg} and \texttt{Local GC} experience substantial fluctuations (exceeding 20\% variation), whereas \texttt{Global GC} maintains a more stable trajectory and achieves earlier performance gains. Similarly, Table~\ref{tab:trainin_dynamics_stats} presents statistical properties of the first-order difference of Figure~\ref{fig:local-global-gc-performance}. As observed, compared to \texttt{Local GC}, \texttt{Global GC} exhibits a lower standard deviation and a higher minimum value, indicating that its fluctuations are significantly reduced throughout training.

\begin{table}[htbp]
    \centering
\caption{Statistical analysis of training dynamics (first-order accuracy difference), showing mean, standard deviation, minimum values, and qualitative performance and fluctuation levels.}
    \scriptsize
    \resizebox{0.5\linewidth}{!}{
    \begin{tabular}{cccccc}
        \toprule
         Model& Mean & Std. & Min & Performance & Fluctuation\\
         \midrule
         FedAvg &  0.052 & 7.33 &  -23.5 &Low &High\\
         Local GC&  0.065 & 6.18 &  -20.1 &\textbf{High} &High\\
         Global GC& 0.062 & 3.70 &  -17.3 &Medium &\textbf{Low}\\
         \midrule
         GC-Fed   & 0.067 & 4.33 &  -18.2 &\textbf{High} &\textbf{Low}\\
         \bottomrule
    \end{tabular}
    }
    \vspace{-3mm}    
    
    \label{tab:trainin_dynamics_stats}
\end{table}

\begin{algorithm}[ht]
\DontPrintSemicolon
\caption{GC-Fed}
\label{alg:GC-FED}

\Input{$N$, $C$, $E$, $R$, $\eta$, $\mathbf{w}_0$, $\lambda$, 
       total layers $L$}
\Output{Global parameters $\mathbf{w}_T$}

\For(\tcp*[h]{Communication rounds}){$t \gets 0$ \KwTo $R$}{

    Randomly select a set of clients $\mathcal{K}$ of size $C \cdot N$\;
    
    \For(\tcp*[h]{In parallel}){$k \in \mathcal{K}$}{
        $\mathbf{w}_{k}^t \gets \mathbf{w}^{t}$\;
        \For(\tcp*[h]{Local epochs}){$e \gets 1$ \KwTo $E$}{
            $\mathbf{G}^{t} \gets \nabla F_k(\mathbf{w}_{k}^{t})$\;
            \For(\tcp*[h]{Layers}){$l \gets 1$ \KwTo $L$}{\label{code:localgc-beginning}
                $L_{\text{local}} \gets \lfloor \lambda \cdot L \rfloor$\;
                \If{$l \le L_{\text{local}}$}{
                    $\Tilde{\mathbf{G}}_k^{t}[l] \gets \mathbf{P}\,\mathbf{G}_k^{t}[l]$ \tcp*[h]{Local GC};
                }
            }\label{code:localgc-end}
            $\mathbf{w}_{k}^{t} \gets \mathbf{w}_{k}^{t} - \eta\,\Tilde{\mathbf{G}}_k^{t}$\
        }
        $\Delta_k^{t+1} \gets \mathbf{w}_{k}^{t} - \mathbf{w}^t$\;
    }
    $\Delta^{t+1} \gets \frac{1}{|\mathcal{K}|} \sum_{k \in \mathcal{K}} \Delta_k^{t+1}$\;
    $\Tilde{\Delta}^{t+1} \gets \mathbf{P}\,\Delta^{t+1}$ \tcp*[h]{Global GC}\;\label{code:globalgc} 
    $\mathbf{w}^{t+1} \gets \mathbf{w}^t + \Tilde{\Delta}^{t+1}$\;
}

\end{algorithm}

Building on these insights, we propose \texttt{GC-Fed}, a hybrid approach that integrates the stability and rapid convergence of \texttt{Global GC} with the superior performance of \texttt{Local GC}. A direct simultaneous application of both methods is ineffective, as locally centralized updates lie on a zero-mean hyperplane, rendering global centralization redundant. To address this, we exclude the classifier layer from \texttt{Local GC} and apply GC only during global aggregation. Furthermore, to enhance flexibility, the target classifier layers can be dynamically adjusted via the parameter $\lambda$, allowing adaptive control over the centralization process. As shown in Figure~\ref{fig:local-global-gc-performance} and Table~\ref{tab:trainin_dynamics_stats}, \texttt{GC-Fed} achieves the best of both worlds, offering both stability and high performance.

Our proposed algorithm, \texttt{GC-Fed}, is detailed in Algorithm~\ref{alg:GC-FED}. Similar to \texttt{FedAvg}, \texttt{GC-Fed} begins by selecting a subset of clients to participate in each communication round, ensuring a realistic FL setting. Each selected client initializes its local model using the global model received from the server. During local training, clients compute gradients using mini-batch SGD. Before updating the local model parameters, the selected layers are first projected (line \ref{code:localgc-beginning}-\ref{code:localgc-end}), where the number of layers to be centralized is determined as $L_{\text{local}} \gets \lfloor \lambda \cdot L \rfloor$. Subsequently, the local weights are updated. For most of our experiments, we explicitly select the last fully-connected layer for the MLP, CNN, and ResNet18 models and the last three fully-connected layers for VGG11 as a borderline. We further investigate variations in layer selection in Section~\ref{sec:layer-borderline}.

Once local training is complete, the server aggregates all local updates ($\Delta_k$) and applies a second projection before computing the final global model update (line~\ref{code:globalgc}). Importantly, the global projection does not require explicit layer specification, as the locally projected layers remain unaffected by this step. This ensures consistency across local and global updates while preserving the structural integrity of the model.

As shown in the algorithm, our approach requires no additional communication, and neither the server nor the clients need extra storage for exchanging reference points to correct gradients during local training or global aggregation. Furthermore, the GC computation is highly efficient, introducing minimal overhead \cite{yong2020gradient}.

\subsection{Theoretical Analysis}

Here, we show how projection can approach the optimum point faster than \texttt{FedAvg} algorithm by comparing the one-step gradient of \texttt{FedAvg} and \texttt{GC-Fed}.
We first introduce a few related notations beforehand as
$\overline{\mathbf{G}}_\tau=\sum_{k=1}^N p_k \nabla F_k(\mathbf{w}_k^\tau) $, thus $\mathbb{E}[\mathbf{G}_\tau]=\overline{\mathbf{G}}_\tau$, similarly,  $\mathbb{E}[\Tilde{\mathbf{G}}_\tau]=\overline{\Tilde{\mathbf{G}}}_\tau$.
Also, the optimum point $\mathbf{w}^*=\arg \min F(\mathbf{w})$ can be decomposed as follows. 
\begin{equation}\label{po}
\mathbf{w}^* = \mathbf{w}^*_{\parallel} + \mathbf{w}^*_{\perp},
\end{equation}
where $\mathbf{w}^*_{\parallel} = \mathbf{P} \mathbf{w}^*$ is the component in the subspace and $\mathbf{w}^*_{\perp} = (\mathbf{I} - \mathbf{P}) \mathbf{w}^*$ is the component orthogonal to subspace.
As an intermediate step to the main theorem, we have an assumption for the optimum point condition and a lemma for one step update as follows.
\begin{assumption}\label{AS_ortho}
\textit{The orthogonal component of $\mathbf{w}^*$ to subspace is zero: $(\mathbf{I} - \mathbf{P}) \mathbf{w}^*=0$.}
\end{assumption}

\begin{lemma}\label{lemma1}
    One step update of Projected Gradient reduces the gap of $\mathbf{w}^*$ and $\mathbf{w}^t$, by $\eta_\tau^2\|\overline{\Tilde{\mathbf{G}}}_\tau\|^2 + \eta_\tau^2 \| \Tilde{\mathbf{G}}_\tau - \overline{\Tilde{\mathbf{G}}}_\tau \|^2 -2\eta_\tau \langle\overline{\mathbf{w}}_\tau - \mathbf{w}^*, \overline{\Tilde{\mathbf{G}}}_\tau \rangle$.
\end{lemma}

\begin{theorem}\label{th1}
    Let assumption \ref{AS_ortho} hold, then from Lemma \ref{lemma1} , it follows that   \texttt{GC-Fed} yields smaller gap than \texttt{FedAvg} by $\eta_\tau^2\|\mathbf{e}^\intercal \mathbf{G}_\tau \|^2+\eta_\tau^2 \|\mathbf{e}^\intercal( \mathbf{G}_\tau - \overline{\mathbf{G}}_\tau) \|^2$.
\end{theorem}

It suggests that the benefit of projection comes from the variance reduction such that the gap is reduced by nonnegative values $\eta_\tau^2\|\mathbf{e}^\intercal \mathbf{G}_\tau \|^2$ and $\|\mathbf{e}^\intercal( \mathbf{G}_\tau - \overline{\mathbf{G}}_\tau) \|^2$.
The proofs of \textbf{Lemma} \ref{lemma1} and \textbf{Theorem} \ref{th1} can be found in Appendix B.

\section{Experiment}

\begin{table*}[ht]
\caption{Summary of Experimental Resources}
\scriptsize
\centering
\resizebox{1.0\linewidth}{!}{%
\renewcommand{\arraystretch}{0.5} % Adjust row height if needed
\begin{tabular}{@{}c c p{11cm}@{}}
\toprule
Type& Name &Description\\
\midrule
\multirow[c]{9}{*}{Data} 
  & EMNIST~\cite{cohen2017emnist}       & 62 classes with 697,932 training data and 116,323 test data, 28$\times$28 pixels in grayscale \\
\cmidrule{2-3}
  & CIFAR10~\cite{Krizhevsky09learningmultiple}      & 10 classes with 50,000 training data and 10,000 test data, 32$\times$32 pixels in RGB color \\
\cmidrule{2-3}
  & CIFAR100~\cite{Krizhevsky09learningmultiple}     & 100 classes with 50,000 training data and 10,000 test data, 32$\times$32 pixels in RGB color \\
\cmidrule{2-3}
  & TinyImageNet~\cite{le2015tiny} & 200 classes with 100,000 training data and 10,000 test data, 64$\times$64 pixels in RGB color \\
% \cmidrule{2-3}
%   & PathMNIST~\cite{medmnistv1}    & 9 classes with 89,996 training data and 7,180 test data, 28$\times$28 pixels in grayscale \\
% \cmidrule{2-3}
%   & TissueMNIST~\cite{medmnistv1}  & 8 classes with 165,466 training data and 47,280 test data, 28$\times$28 pixels in grayscale \\
% \cmidrule{2-3}
%   & OrganAMNIST~\cite{medmnistv1}  & 11 classes with 34,561 training data and 17,778 test data, 28$\times$28 pixels in grayscale \\
\midrule

\multirow[c]{16}{*}{Model} 
  & MLP      & Custom model with three fully connected layers \\
\cmidrule{2-3}
  & CNN      & Two convoluation layers and two fully connected layers used in~\cite{mcmahan2017communication} \\
\cmidrule{2-3}
  & VGG11~\cite{simonyan2014very} & 11 layers in total (8 convolutional and 3 fully connected). The final layer is adjusted based on the number of classes. Implemented using the PyTorch~\cite{paszke2019pytorch} version.\\

\cmidrule{2-3}
& ResNet18~\cite{he2016deep} & 18 layers in total (1 initial convolution layer, 16 convolutional layers in residual blocks, and 1 fully connected layer). The final layer is adjusted based on the number of classes. Implemented using the PyTorch version. The first convolution layer's kernel size is reduced to 3×3 for small-scale images, and the max pooling layer has been removed. Batchnorm is replaced with Groupnorm.\\
\midrule

\multirow[c]{44}{*}{FL Algorithm} 
  & FedAvg~\cite{mcmahan2017communication}     
      & Fundamental FL algorithm (historically independent).\\
\cmidrule{2-3}
  & FedProx~\cite{li2020federated}    
      & Introducing a proximal term (L2 difference between global and local model) in the local training step, keeping local and global models close (historically dependent).\\
\cmidrule{2-3}
  & SCAFFOLD~\cite{karimireddy2020scaffold}   
      & Introducing local and global control variates to correct local gradient directions while accounting for global updates (historically dependent).\\
\cmidrule{2-3}
  & FedDyn~\cite{acar2021federated}     
      & Introducing a dynamic regularization term that aligns each client’s objective with the global optimum (historically dependent).\\
\cmidrule{2-3}
  & FedNTD~\cite{lee2022preservation}     
      & Introducing a proximal loss (KL divergence between global and local logits, except for the true class) to preserve knowledge about out-of-distribution classes (historically dependent).\\
\cmidrule{2-3}
  & FedVARP~\cite{jhunjhunwala2022fedvarp}    
      & Introducing global optimization inspired by SAGA~\cite{defazio2014saga}, retaining all clients’ local updates on the server to adjust the update direction during aggregation (historically dependent).\\
\cmidrule{2-3}
  & FedLC~\cite{mendieta2022local}      
      & Introducing logit calibration to mitigate prediction bias caused by missing classes in local datasets (historically independent).\\
\cmidrule{2-3}
  & FedDecorr~\cite{shi2022towards}  
      & Introducing a decorrelation regularization that minimizes the Frobenius norm of inter-dimensional correlation to alleviate dimensional collapse (historically independent).\\
\cmidrule{2-3}
  & FedSOL~\cite{lee2024fedsol}
      & Introducing an orthogonal learning approach combining a KL-based proximal objective with SAM~\cite{foret2020sharpness}-like weight perturbation (historically dependent).\\
\cmidrule{2-3}
  & FedACG~\cite{kim2024communication}     
      & Introducing global optimization with a look-ahead mechanism to anticipate and correct client update directions during aggregation (historically dependent).\\
\bottomrule

\end{tabular}
}

\label{tab:summary}
\end{table*}

\subsection{Experimental Setup}

\paragraph{\textbf{Datasets, Models, and Baselines.}} We evaluate our approach on EMNIST~\cite{cohen2017emnist}, CIFAR-10, CIFAR-100~\cite{Krizhevsky09learningmultiple}, and TinyImageNet~\cite{le2015tiny}. Standard augmentation (random cropping and horizontal flipping) is applied only during training. We experiment with a variety of architectures, from simple MLPs and CNNs to larger models such as VGG11 and ResNet18. For benchmarking, we compare against 10 state-of-the-art baselines published up to 2024. A detailed summary of the experimental resources is provided in Table~\ref{tab:summary}.

\begin{figure}[htbp]
    \centering
    \includegraphics[width=\linewidth]{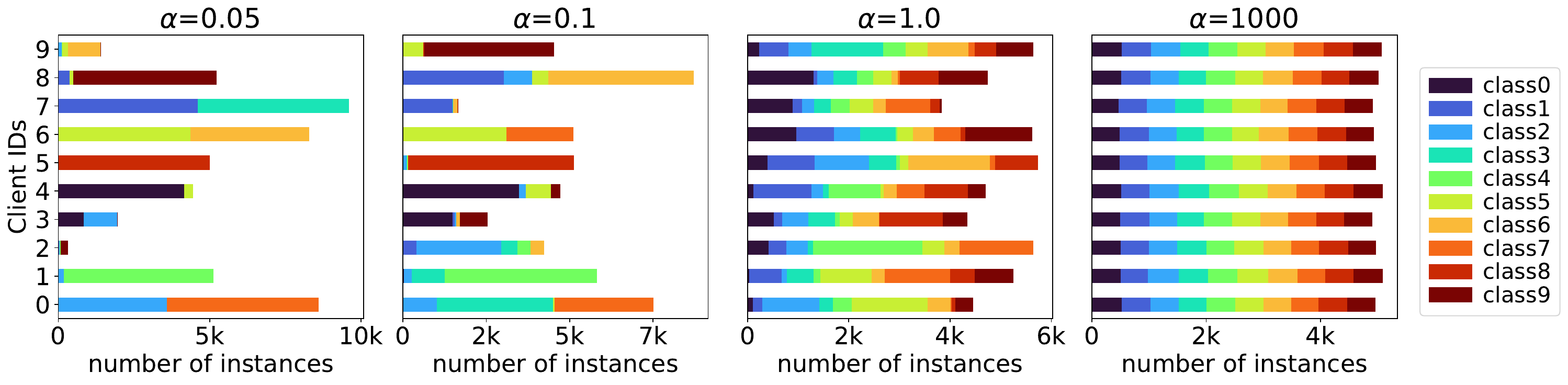}
    \caption{Data partitioning across different LDA $\alpha$ values in a 10-class, 10-client setting. Lower $\alpha$ values induce higher heterogeneity in class distribution and data volume per client, while higher $\alpha$ values yield a more homogeneous distribution.}
    \label{fig:data-distribution}
\end{figure}

\paragraph{\textbf{Training and Evaluation.}} In each communication round, we randomly select $K$ clients without replacement to simulate partial participation in a \textit{non-i.i.d.} setting. Each client trains locally for 5 epochs with a batch size of 50 using SGD (learning rate 0.01, momentum 0.9, weight decay $\num{1e-5}$), without a learning rate scheduler. The cross-entropy loss is computed locally and aggregated to update the global model. Test accuracy is evaluated on a held-out set after every aggregation round. Data partitioning follows an unbalanced latent Dirichlet allocation (LDA) scheme; its impact of parameter $\alpha$ is illustrated in Figure~\ref{fig:data-distribution}.

\subsection{Experimental Results}
\subsubsection{Baseline Algorithm Validation}

\begin{table}[htbp]
\caption{Comparison of \textbf{top-1 test accuracy (\%)} of FL algorithms under weak heterogeneous and homogeneous settings with moderate partial participation (CIFAR-10, CNN, R: 800). Bold text indicates the highest accuracy in each column for baseline and our methods respectively, while underlined text represents the highest accuracy. Values in parentheses denote the standard deviation across different seeds.}

\scriptsize
\centering
\resizebox{0.5\linewidth}{!}{
\begin{tabular}{ccccc}
\toprule
\multicolumn{1}{c}{LDA $\alpha$} &\multicolumn{2}{c}{1.0} &\multicolumn{2}{c}{1000} \\
\multicolumn{1}{c}{Participation}  &5/100&10/100&5/100&10/100\\
\midrule
\texttt{FedAvg}~\cite{mcmahan2017communication}     & 73.50 (1.97) & 74.44 (1.69) & 76.88 (0.76) & 77.55 (0.77) \\
\midrule
\texttt{FedProx}~\cite{li2020federated}   & 74.31 (1.28) & 75.62 (1.47) & 77.58 (0.30) & 77.94 (1.00) \\
\texttt{Scaffold}~\cite{karimireddy2020scaffold}  & 74.34 (2.74) & 77.18 (1.18) & \textbf{79.47} (1.27) & 78.57 (0.88) \\
\texttt{FedDyn}~\cite{acar2021federated}     & \textbf{77.03} (0.77) & \textbf{\underline{79.45}} (0.49)  & 79.04 (0.03) & \textbf{80.28} (0.01) \\
\texttt{FedNTD}~\cite{lee2022preservation}     & 75.18 (0.56) & 75.33 (0.59)  & 77.03 (0.36) & 76.76 (0.70) \\
\texttt{FedVARP}~\cite{jhunjhunwala2022fedvarp}    & 74.50 (1.45) & 74.80 (1.19)  & 76.72 (1.11) & 77.07 (0.65) \\
\texttt{FedLC}~\cite{zhang2022federated}     & 73.49 (1.00) & 74.98 (0.77) & 77.44 (1.11) & 77.50 (0.96) \\
\texttt{FedDecorr}~\cite{shi2022towards}     & 72.28 (1.89) & 74.09 (1.63) & 76.11 (0.68) & 76.32 (0.78) \\
\texttt{FedSOL}~\cite{lee2024fedsol}     & 73.91 (2.18) & 74.60 (1.41) & 77.39 (1.35) & 77.06 (2.05) \\
\texttt{FedACG}~\cite{kim2024communication}     & 72.16 (1.03) & 73.63 (1.97) & 77.37 (1.06) & 77.83 (0.43) \\
\midrule
\texttt{LocalGC} & 78.68 (0.02) & \textbf{79.29} (0.33) & \textbf{\underline{80.88}} (0.53) & \textbf{\underline{80.92}} (0.34) \\
\texttt{GlobalGC} & 76.55 (0.26) & 77.05 (0.71) & 78.56 (0.50) & 79.14 (0.52) \\
\texttt{GC-Fed}   & \textbf{\underline{78.73}} (0.78) & 79.25 (0.46) & 80.20 (0.39) & 80.56 (0.30) \\
\bottomrule
\end{tabular}

}\label{tab:main-result2}
\end{table}

We begin our evaluation by assessing the performance of our baseline algorithm on the CIFAR-10 dataset using a convolutional neural network (CNN) model. Experiments were conducted under both moderately \textit{non-i.i.d.} conditions ($\alpha = 1.0$) and \textit{i.i.d.} distributions ($\alpha = 1000$). As shown in Table~\ref{tab:main-result2}, most algorithms achieve performance comparable to or surpassing that of \texttt{FedAvg}, with performance discrepancies remaining within approximately 5\%. This result confirms the robustness of our implementation and the reliability of our experimental setup for evaluating the baseline algorithms. Although GC-based methods consistently outperform other approaches across all scenarios, we note that this dataset distribution is not representative of real-world FL settings and is not the primary focus of our study.

\subsubsection{Small Scale Experiment}

\begin{table*}[htbp]
\caption{Comparison of top-1 test accuracy (\%) of FL algorithms across three datasets under a \textit{non-i.i.d.} setting. EMNIST is evaluated using an MLP for 200 rounds, while CIFAR-10 and CIFAR-100 are evaluated using a CNN for 800 rounds.}

\label{tab:small-scale-result}

\scriptsize
\centering

\resizebox{1.0\linewidth}{!}{
\begin{tabular}{cccccccccc}
\toprule
\multicolumn{1}{c}{Dataset}   & EMNIST       & \multicolumn{4}{c}{CIFAR-10}  & \multicolumn{4}{c}{CIFAR-100}  \\

\multicolumn{1}{c}{Participation} & 5/1000  & \multicolumn{2}{c}{5/100}         & \multicolumn{2}{c}{5/200}         & \multicolumn{2}{c}{5/100}           & \multicolumn{2}{c}{5/200} \\

\multicolumn{1}{c}{LDA $\alpha$} & 0.05    & 0.05        & 0.10          & 0.05         & 0.10          & 0.05         & 0.10          & 0.05         & 0.10 \\
\midrule
\texttt{FedAvg}~\cite{mcmahan2017communication}    & 59.45 (3.72)  & 53.74 (3.37)  & 66.44 (1.40)  & 45.14 (4.44)  & 58.05 (3.18)  & 31.88 (0.76)  & 34.20 (0.60)  & 30.92 (1.67)  & 32.36 (0.86) \\
\midrule
\texttt{FedProx}~\cite{li2020federated}   & 60.45 (3.00)  & 53.93 (0.87)  & 67.19 (2.67)  & 49.12 (3.86)  & 60.73 (3.77)  & 32.91 (0.72)  & 35.12 (0.21)  & 30.54 (0.97)  & 33.00 (1.11) \\
\texttt{Scaffold}~\cite{karimireddy2020scaffold}  & 62.52 (3.89)  & 10.00 (Failed)& 10.00 (Failed)& 10.00 (Failed)& 10.00 (Failed)& \textbf{36.41} (1.99)  & \textbf{39.69} (1.94)  & \textbf{35.07} (0.78)  & \textbf{37.83} (1.70) \\
\texttt{FedDyn}~\cite{acar2021federated}    & 41.05 (3.73)  & 10.00 (Failed)& 39.87 (11.13)& 10.00 (Failed)& 22.48 (6.40)& 32.49 (0.90)  & 34.63 (0.73)  & 10.94 (0.80)  & 15.27 (0.23) \\
\texttt{FedNTD}~\cite{lee2022preservation}    & 64.09 (2.18)  & 56.08 (4.50)  & 65.62 (2.31)  & 55.43 (2.26)  & 60.85 (3.79)  & 32.56 (1.67)  & 35.31 (2.06)  & 32.48 (1.52)  & 33.66 (1.12) \\
\texttt{FedVARP}~\cite{jhunjhunwala2022fedvarp}    & \textbf{66.18} (2.62)  & \textbf{62.99} (2.28)  & \textbf{67.69} (1.60)  & 49.66 (7.12)  & \textbf{65.31} (3.70)  & 34.40 (1.13)  & 37.13 (1.01)  & 33.88 (0.75)  & 34.82 (1.67) \\
\texttt{FedLC}~\cite{zhang2022federated}     & 63.60 (2.60)  & 58.05 (0.26)  & 65.95 (1.32)  & \textbf{56.86} (2.16)  & 61.26 (1.83)  & 30.84 (1.88)  & 34.62 (1.13)  & 29.59 (0.52)  & 32.23 (0.19) \\
\texttt{FedDecorr}~\cite{shi2022towards}     & 60.29 (0.70)  & 44.79 (3.02)  & 62.84 (4.09)  & 37.60 (3.25)  & 53.32 (5.09)  & 30.45 (0.87)  & 33.36 (0.54)  & 28.62 (1.04)  & 31.24 (1.46) \\
\texttt{FedSOL}~\cite{lee2024fedsol}    & 62.89 (3.64)  & 58.99 (3.74)  & 65.41 (1.42)  & 52.59 (4.44)  & 61.23 (3.08)  & 32.13 (0.35)  & 34.95 (1.03)  & 30.34 (1.60)  & 31.64 (1.29) \\
\texttt{FedACG}~\cite{kim2024communication}& 58.08 (0.32)  & 53.32 (1.83)  & 62.61 (0.19)  & 42.52 (1.43)  & 59.85 (3.45)  & 34.62 (0.58)  & 35.25 (0.76)  & 32.88 (1.29)  & 34.32 (1.18) \\
\midrule
\texttt{LocalGC} & 63.73 (0.41)  & 64.59 (4.47)  & 71.81 (1.70)  & 62.00 (5.26)  & \textbf{\underline{69.91 (1.70)}}  & 38.43 (0.29)  & 40.52 (0.90)  & 38.21 (1.17)  & \textbf{\underline{40.02}} (0.12) \\
\texttt{GlobalGC} & \textbf{\underline{67.42}} (0.86)  & 10.00 (Failed)& 65.88 (1.32)  & 60.59 (2.02)  & 64.97 (0.70)  & 34.55 (0.54)  & 36.89 (0.71)  & 33.85 (0.92)  & 36.43 (0.57) \\
\textbf{\texttt{GC-Fed}} & 66.63 (1.48)  & \textbf{\underline{66.47}} (2.49)  & \textbf{\underline{72.44}} (2.14)  & \textbf{\underline{64.19}} (2.69)  & 69.91 (0.51)  & \textbf{\underline{39.18}} (0.53)  & \textbf{\underline{41.06}} (0.69)  & \textbf{\underline{39.09}} (0.84)  & 39.99 (0.81) \\
\bottomrule
\end{tabular}

}

\end{table*}

In our small-scale experiments, we evaluate our approach using three different datasets: EMNIST, CIFAR-10, and CIFAR-100. For the EMNIST dataset, we employ an MLP model, while for CIFAR-10 and CIFAR-100, we utilize a CNN. The \textit{non-i.i.d.} condition is severe, with the LDA parameter $\alpha$ ranging from 0.05 to 0.1 (see Figure~\ref{fig:data-distribution}). Additionally, we test with a large number of clients: 1000 for EMNIST and 200 for CIFAR-10 and CIFAR-100.

As shown in Table~\ref{tab:small-scale-result}, our proposed \texttt{GC-Fed} consistently outperforms standard baselines across various datasets and model architectures. For instance, on the EMNIST dataset using an MLP (with 5/1000 participation and LDA $\alpha=0.05$), \texttt{GC-Fed} achieves 66.63\%, which is a +7.18 percentage point improvement over \texttt{FedAvg} (59.45\%) and slightly surpasses \texttt{FedVARP} (66.18\%). In the CNN experiments on CIFAR-10, the performance gains over \texttt{FedAvg} range from approximately +6.0 percentage points (e.g., 72.44\% vs. 66.44\% for 5/100 participation with $\alpha=0.10$) to as high as +19.1 percentage points (64.19\% vs. 45.14\% for 5/200 participation with $\alpha=0.05$). When compared against the best performing non–GC baseline (e.g., \texttt{FedVARP} or \texttt{FedLC}), the improvements vary between +3.5 and +7.3 percentage points. For CIFAR-100 with CNN, the improvements over \texttt{FedAvg} are in the range of approximately +6.9 to +8.2 percentage points, while the gains over the best non–GC baseline are more modest, spanning roughly +1.4 to +2.8 percentage points.

Overall, these results demonstrate that \texttt{GC-Fed} not only provides substantial improvements over \texttt{FedAvg}—with gains ranging from +6.0 to +19.1 percentage points—but also consistently outperforms the best non–GC-based approaches by +0.5 to +7.3 percentage points.

Focusing on the CIFAR-10 results, we observe a consistent trend: as the LDA parameter $\alpha$ decreases, the model performance deteriorates. This pattern holds across all algorithms. Furthermore, as the participation ratio decreases (i.e., with an increasing number of total clients), the performance also declines. Notably, several algorithms, including \texttt{FedDyn} and \texttt{SCAFFOLD}, fail to converge in the CIFAR-10 experiments, a trend reported in multiple studies~\cite{lee2022preservation,varno2022adabest,lee2024fedsol,reddi2021adaptive,jhunjhunwala2022fedvarp,baumgart2024not}. Specifically, \texttt{Global GC} fails to train under $\alpha=0.05$ and a 5/100 participation ratio. This failure occurs immediately after the system encounters multiple clients with only a single class label. Given that CIFAR-10 comprises only 10 classes, the current \textit{non-i.i.d.} setup results in some clients containing only one class, which negatively impacts the overall model, primarily due to gradient explosion during local training~\cite{crawshaw2023episode}. While careful hyperparameter tuning or additional techniques such as gradient clipping~\cite{mikolov2012statistical} could mitigate this issue, we do not apply such adjustments in this study so as to isolate the core performance factors.

Furthermore, experiments on EMNIST with an MLP model confirm the effectiveness of GC, even for fully connected layers. This observation suggests that the impact of GC is not confined to convolutional layers and supports our hyperplane-based reference perspective. Consequently, GC can be extended to various architectures, including transformer-based models~\cite{vaswani2017attention,dosovitskiy2021an}.

\subsubsection{Large Scale Experiment}

\begin{table*}[htbp]
\caption{Comparison of top-1 test accuracy (\%) on CIFAR-100 and TinyImageNet using VGG11 and ResNet18. Results are smoothed over the last 10 rounds and averaged across three independent runs. \texttt{SCAFFOLD}$^\dag$, \texttt{FedDyn}$^\dag$, and \texttt{FedVARP}$^\dag$ with VGG11 are excluded due to hardware limitations.}
\scriptsize
\centering
\resizebox{1.0\linewidth}{!}{
\begin{tabular}{cccccccccc}
\toprule

\multirow{3}{*}{Model}&Data & \multicolumn{4}{c}{CIFAR-100} & \multicolumn{4}{c}{TinyImageNet} \\
% \cmidrule(lr){2-5} \cmidrule(lr){6-9}
&Participation& \multicolumn{2}{c}{5/400} &\multicolumn{2}{c}{5/800}&\multicolumn{2}{c}{5/500}&\multicolumn{2}{c}{5/1000} \\
 &LDA& 0.05 & 0.1 & 0.05 & 0.1& 0.05 & 0.1 & 0.05 & 0.1 \\
\midrule

&\texttt{FedAvg}~\cite{mcmahan2017communication} & 24.55 (2.22) & 32.38 (2.36) & 16.53 (0.02)& 25.97 (2.24)& 24.86 (0.88) & 29.01 (1.12)& 18.28 (0.32)& 24.65 (0.58)  \\
\cmidrule(lr){2-10} 
&\texttt{FedProx}~\cite{li2020federated}  & 24.87 (2.08)& 32.54 (2.45)& 16.57 (0.13)& 25.90 (1.83)&24.95 (0.49)&28.93 (0.47)&18.44 (0.36)&24.77 (0.76)\\
&\texttt{SCAFFOLD}$^\dag$~\cite{karimireddy2020scaffold} & N/A&N/A&N/A&N/A&N/A&N/A&N/A&N/A\\
&\texttt{FedDyn}$^\dag$~\cite{acar2021federated} & N/A&N/A&N/A&N/A&N/A&N/A&N/A&N/A\\
&\texttt{FedNTD}~\cite{lee2022preservation}   & 26.46 (0.74)& \textbf{36.62} (0.21)& 18.33 (0.12)& \textbf{30.08} (1.02)&27.96 (1.08)&\textbf{33.39} (0.25)& 19.29 (1.14)&\textbf{26.94} (0.25)\\

&\texttt{FedVARP}$^\dag$~\cite{jhunjhunwala2022fedvarp} & N/A&N/A&N/A&N/A&N/A&N/A&N/A&N/A\\
VGG11&\texttt{FedLC}~\cite{zhang2022federated}  & \textbf{29.56 }(0.53)& 35.40 (0.84)& \textbf{23.93} (0.56)& 27.75 (0.18)& \textbf{28.26} (0.05)&30.33 (1.23)& \textbf{22.78} (0.66)&22.87 (0.92)\\

&\texttt{FedDecorr}~\cite{shi2022towards} &23.95 (1.96)&31.86 (1.31)&16.05 (0.01)&25.51 (2.65)&23.73 (0.50) & 28.53 (0.41)& 17.56 (0.87) &23.94 (0.36)\\
&\texttt{FedSOL}~\cite{lee2024fedsol}   & 27.16 (1.32)& 33.35 (1.43)& 19.14 (0.37)& 27.40 (1.56) &24.90 (0.39)&29.01 (0.40)&19.19 (0.17)&24.62 (0.49)\\

&\texttt{FedACG}~\cite{kim2024communication}  &19.91 (0.18) &29.01 (1.91) & 8.49 (0.60)&15.97 (1.74) &13.83 (0.09)&24.72 (0.01)&6.99 (0.45)&13.59 (1.01)\\
\cmidrule(lr){2-10} 
&\texttt{LocalGC}  & 36.71 (2.36)& 41.48 (1.70)& 29.39 (0.01)& 34.27 (1.06)& 33.54 (0.09)&35.91 (0.35)&28.73 (0.12)&30.88 (0.79)\\
&\texttt{GlobalGC} & 31.18 (1.04)&37.51 (1.50)&26.41 (0.70)& 31.71 (0.76)&28.92 (0.73)&32.09 (0.29)& 24.95 (0.22) &26.61 (0.56)\\
&\textbf{\texttt{GC-Fed}}  &\textbf{\underline{37.03}} (0.68)& \textbf{\underline{42.37}} (1.27)&\textbf{\underline{31.18}} (0.31)&\textbf{\underline{35.27}} (1.06)& \textbf{\underline{33.97}} (0.65)&\textbf{\underline{36.13}} (1.18)&\textbf{\underline{29.54}} (0.07)&\textbf{\underline{31.24}} (0.21)\\

\bottomrule
\toprule

&\texttt{FedAvg}~\cite{mcmahan2017communication} & 13.61 (1.69) & 23.03 (2.41)& 9.91 (0.01) & 18.20 (1.57) & 13.38 (1.25)& 24.10 (0.09)& 7.46 (0.03)& 15.24 (0.82)  \\
\cmidrule(lr){2-10} 
&\texttt{FedProx}~\cite{li2020federated}  & 14.10 (1.92) & 23.55 (2.25) & 10.05 (0.01)& 18.54 (1.74)& 14.13 (0.99)& 24.44 (0.41) & 7.72 (0.11)& 15.44 (0.78)\\
&\texttt{SCAFFOLD}~\cite{karimireddy2020scaffold} &25.33 (1.45)&\textbf{37.41} (1.84)&18.37 (0.92) & \textbf{26.20} (1.92)&\textbf{24.21} (0.98)& \textbf{33.81} (0.47)& 13.90 (0.99)&\textbf{23.68} (0.56)\\
&\texttt{FedDyn}~\cite{acar2021federated} &7.78 (1.52)&17.56 (3.26)&6.57 (0.55)&9.82 (1.10)&3.32 (0.64) & 12.77 (1.05) & 0.50 (Failed) & 3.47 (0.61)\\
&\texttt{FedNTD}~\cite{lee2022preservation}   & 15.99 (1.43)&27.74 (0.72) & 11.03 (1.51)& 21.66 (1.11)&  15.63 (0.57)& 26.86 (0.50) & 8.46 (0.05)&18.07 (0.68)\\
&\texttt{FedVARP}~\cite{jhunjhunwala2022fedvarp} & 21.91 (0.58)& 32.57 (1.47) & 16.60 (0.94) & 24.56 (2.38)&  23.22 (1.73)&31.77 (0.65) & 14.33 (1.11) & 22.67 (1.40)\\
ResNet18&\texttt{FedLC}~\cite{zhang2022federated}  & \textbf{26.02} (1.38)& 33.27 (1.17)& \textbf{19.68} (0.96)& 25.51 (1.14) &  23.03 (0.76)& 29.89 (0.56) & \textbf{15.43} (0.71)& 21.87 (0.23)\\

&\texttt{FedDecorr}~\cite{shi2022towards} &16.33 (2.14)&24.72 (1.52)&11.68 (0.80)&19.45 (2.18)&15.90 (1.12)&25.42 (0.92)&10.00 (0.73) &17.03 (0.53)\\
&\texttt{FedSOL}~\cite{lee2024fedsol}   & 19.47 (1.53)& 27.83 (1.13) & 15.05 (0.79)& 22.38 (0.67) & 18.67 (1.38)& 26.25 (0.24) & 11.62 (0.52)&17.23 (0.45)\\
&\texttt{FedACG}~\cite{kim2024communication}  &9.81 (0.93)&16.46 (0.17)&5.35 (0.58)&8.84 (0.55)&18.35 (0.56)&24.89 (0.47)&4.67 (0.61)&10.21 (1.00)\\
\cmidrule(lr){2-10} 
&\texttt{LocalGC}   & 35.90 (3.02)& 42.72 (1.70)& 26.70 (1.79) &34.02 (2.47) &  32.61 (1.85)&36.92 (0.26)  & 24.12 (0.13)& 28.40 (0.41)\\
&\texttt{GlobalGC}   & 24.45 (1.76)& 32.80 (1.59)& 21.56 (0.91)& 28.23 (0.89) &  21.66 (0.53)&29.69 (0.59) & 15.72 (0.84) & 22.90 (0.67)\\
& \textbf{\texttt{GC-Fed}}  & \textbf{\underline{36.54}} (2.80)& \textbf{\underline{43.52}} (1.65)& \textbf{\underline{27.77}} (1.00) & \textbf{\underline{34.57}} (2.03) &  \textbf{\underline{33.18}} (1.97) &\textbf{\underline{37.33}} (0.90) & \textbf{\underline{24.39}} (0.17)& \textbf{\underline{29.45}} (0.32)\\
\bottomrule

\end{tabular}
}

\label{tab:large-scale}
\end{table*}

We conduct experiments on a large-scale model and dataset (TinyImageNet) with a large number of clients to simulate the real-world cross-device scenario. We employ two widely used neural network architectures: ResNet-18~\cite{he2016identity} with 11.7M parameters (44.7~MB) and VGG-11~\cite{simonyan2014very} with 132.8M parameters (506.8~MB).

As we can see in Table~\ref{tab:large-scale}, our proposed \texttt{GC-Fed} outperforms most baselines on this large-scale dataset with larger models. In particular, compared to \texttt{FedAvg}, \texttt{GC-Fed} shows improvements that range from approximately +6.6 to +22.9 percentage points across various configurations. When compared to the best-performing baseline (such as \texttt{FedLC}, \texttt{FedNTD}, or \texttt{SCAFFOLD}, depending on the setting), the improvements range from about +2.7 to +10.5 percentage points. These consistent gains are evident across both CIFAR‑100 and TinyImageNet datasets, as well as across different participation settings (e.g., 5/400, 5/800, 5/500, 5/1000) and LDA values (0.05 and 0.1). Overall, these results underscore the robustness and effectiveness of \texttt{GC-Fed} in enhancing model performance in FL scenarios.

Note that we do not run experiments for \texttt{SCAFFOLD}, \texttt{FedDyn}, and \texttt{FedVARP} on VGG-11 due to hardware limitations. In particular, \texttt{SCAFFOLD} and \texttt{FedDyn} require each client to store previous information (e.g., local control variates), while \texttt{FedVARP} requires the server to retain all clients’ past local updates. For instance, training VGG-11 with \( N \) clients incurs an additional storage cost of \( O(N) \) for each algorithm, which becomes prohibitive as \( N \) grows large. Although \texttt{SCAFFOLD} and \texttt{FedDyn} remain feasible in real-world tests---since the additional information is stored on each client---the storage requirement for \texttt{FedVARP} imposes a substantial burden on the server. Consequently, despite the clustering-based approach proposed in the original \texttt{FedVARP} paper~\cite{jhunjhunwala2022fedvarp}, we do not incorporate it into our study.

Furthermore, \texttt{FedACG} consistently exhibits lower performance compared to \texttt{FedAvg}. In our experiments, the global momentum conflicts with the local momentum. In the original \texttt{FedACG} paper~\cite{kim2024communication}, the local momentum is set to zero, whereas we set it to 0.9, consistent with common practice in FL research. Although careful hyperparameter tuning might mitigate this issue, we do not explore this direction in the present study.

\subsubsection{Training Dynamics}
\begin{figure*}[htbp]
    \centering
    \includegraphics[width=\linewidth]{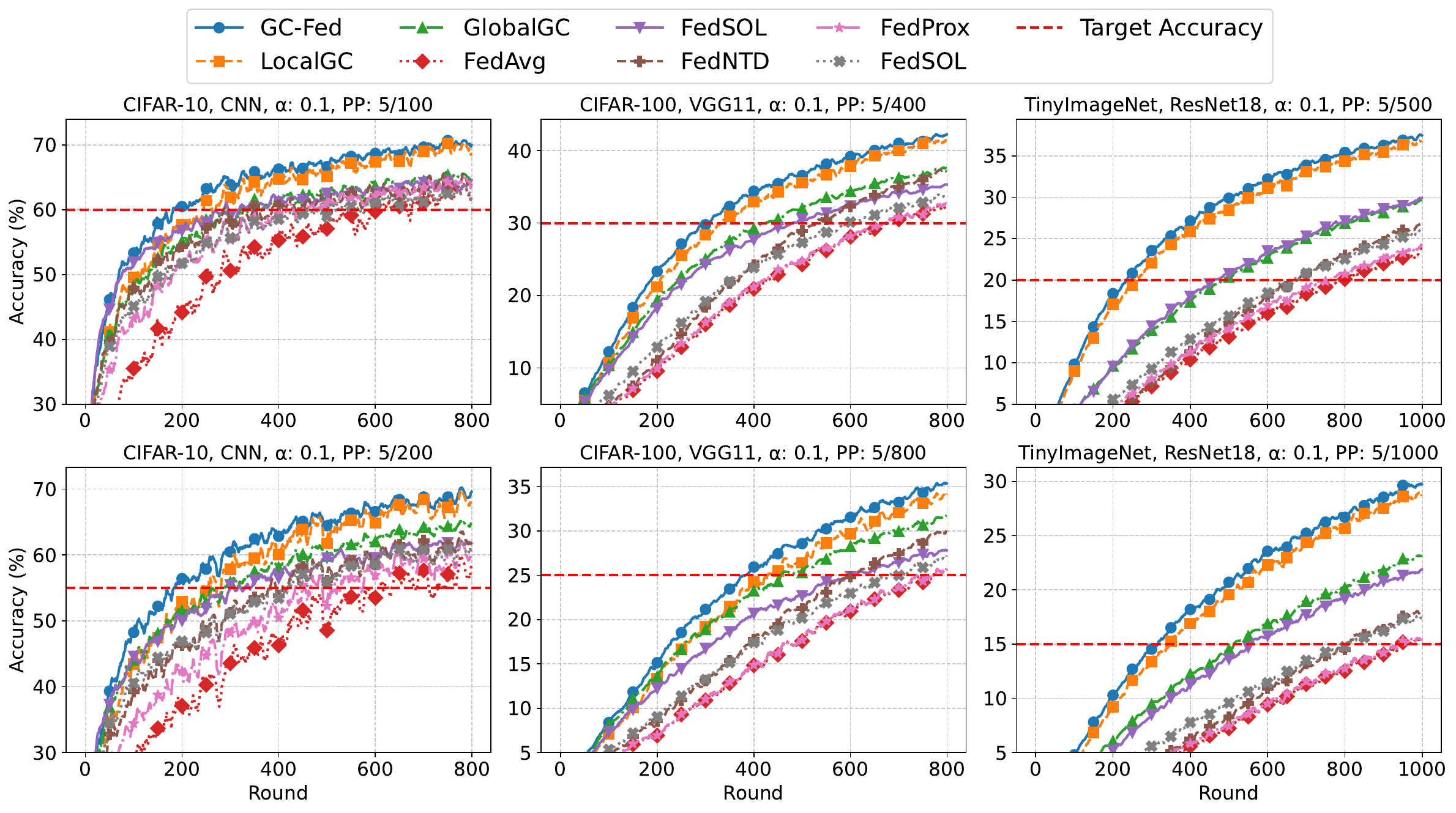}
    \caption{Training dynamics of various FL algorithms on CIFAR-10, CIFAR-100, and TinyImageNet, using CNN, VGG11, and ResNet18 models, respectively. “PP” denotes partial participation. Curves are smoothed via a moving average (window size 10) for enhanced clarity.}
    \label{fig:training-dynamic-supple}
\end{figure*}

In Figure~\ref{fig:training-dynamic-supple}, we present the training dynamics for a subset of algorithms that consistently outperform \texttt{FedAvg} in our experiments. For the CIFAR-10 (CNN, $\alpha=0.1$, PP: 5/100) setting with a target accuracy of 60\%, \texttt{GC-Fed} converges rapidly at round 182, while \texttt{Local GC} and \texttt{Global GC} reach the target at rounds 244 and 322, respectively. In contrast, \texttt{FedAvg} requires 598 rounds to converge, with other baseline methods, such as \texttt{FedLC}, \texttt{FedNTD}, \texttt{FedProx}, and \texttt{FedSOL}, exhibiting convergence rounds between 287 and 425. A similar trend is observed in the CIFAR-100 (VGG11, $\alpha=0.1$, PP: 5/400) experiment, where the target accuracy is set to 30\%. Here, \texttt{GC-Fed} reaches 30\% accuracy at round 309, followed by \texttt{Local GC} and \texttt{Global GC} at rounds 332 and 421, respectively, whereas \texttt{FedAvg} converges only at round 676 and the remaining methods achieve convergence within the range of 479 to 665 rounds.

In the more challenging TinyImageNet (ResNet18, $\alpha=0.1$, PP: 5/500) setting with a target accuracy of 20\%, GC-Fed is the fastest to converge, achieving the target at round 236; \texttt{Local GC} follows at round 257, while \texttt{Global GC} and \texttt{FedAvg} converge at rounds 486 and 781, respectively, with other methods converging within intermediate ranges. Adjusting the communication budget further reinforces these observations: in a CIFAR-10 (CNN, $\alpha=0.1$, PP: 5/200) scenario with a target accuracy of 55\%, \texttt{GC-Fed} converges at round 184 in contrast to \texttt{FedAvg}’s 584 rounds. Similarly, in a CIFAR-100 (VGG11, $\alpha=0.1$, PP: 5/800) setting with a target of 25\%, \texttt{GC-Fed} achieves the target at round 385 compared to 773 rounds for \texttt{FedAvg}, and in a TinyImageNet (ResNet18, $\alpha=0.1$, PP: 5/1000) experiment with a target of 15\%, \texttt{GC-Fed} converges at round 316 while \texttt{FedAvg} requires 943 rounds.

Overall, these results demonstrate that \texttt{GC-Fed} not only attains higher performance but also exhibits a substantially faster convergence rate than \texttt{FedAvg} and other competing methods. The accelerated convergence of \texttt{GC-Fed} translates directly into reduced communication overhead, rendering it particularly advantageous for FL scenarios where rapid model updates and efficiency are paramount. These findings highlight the potential of \texttt{GC-Fed} as a robust alternative in heterogeneous federated environments.

\subsubsection{Integration Test}
\begin{table}[htbp]
\caption{Top-1 test accuracy (\%) on CIFAR-100 with ResNet18 after 800 rounds, comparing the integration of \texttt{Global GC} and \texttt{GC-Fed} with other FL baselines.}
\centering
\small
\resizebox{0.6\linewidth}{!}{
    \begin{tabular}{lcccc}
        \toprule
        Participation & 5/400& 5/400& 5/800& 5/800\\
        LDA $\alpha$& 0.05&0.10& 0.05&0.10\\
        \toprule
        \texttt{FedAvg} & 13.55 (1.75)& 24.01 (1.43)& 9.59 (0.32)& 18.41 (1.36)\\
        
        \texttt{FedAvg} (with \texttt{Global GC}) & 24.40 (1.79)& 33.03 (1.36)&21.68 (0.79) & 28.45 (0.68)\\
        
        \texttt{FedAvg} (with \texttt{GC-Fed}) & 36.21 (3.13)& 43.72 (1.45)& 27.52 (1.25)& 34.95 (1.65)\\
        \midrule
        
        \texttt{FedProx} & 13.90 (2.11)& 24.38 (1.43)& 9.78 (0.24) & 18.86 (1.41)\\
        
        \texttt{FedProx} (with \texttt{Global GC}) & 25.60 (1.83)& 33.01 (1.52)& 21.66 (0.60)& 28.37 (0.76)\\
        
        \texttt{FedProx} (with \texttt{GC-Fed}) & 36.30 (2.95) & 44.08 (1.54)& 27.74 (1.16)& 35.19 (1.78)\\
        \midrule
        \texttt{FedLC} & 25.68 (1.42)& 33.78 (0.66)& 19.87 (0.77) & 25.44 (1.20)\\
        
        \texttt{FedLC} (with \texttt{Global GC}) & 29.90 (2.34) & 39.40 (1.05)& 23.70 (0.51)& 29.76 (2.37)\\
        
        \texttt{FedLC} (with \texttt{GC-Fed}) & \textbf{39.91} (1.61)& \textbf{44.94} (0.50)& \textbf{30.65} (0.85)& \textbf{35.29} (2.69)\\
        \midrule
        FedNTD & 16.38 (1.04)& 27.86 (0.61) & 10.99 (1.55) & 21.85 (0.93)\\
        
        \texttt{FedNTD} (with \texttt{Global GC}) & 17.10 (2.30)& 30.12 (0.76)& 15.57 (1.22)& 24.53 (1.12)\\
        
        \texttt{FedNTD} (with \texttt{GC-Fed}) & 33.39 (1.89)& 42.76 (0.11)& 24.71 (0.61)& 33.94 (1.54)\\
        \midrule
        \texttt{FedDecorr} & 15.86 (2.61)& 25.59 (0.65)& 11.18 (1.30) & 19.97 (1.65)\\
        
        \texttt{FedDecorr} (with \texttt{Global GC}) & 26.55 (1.34)& 34.34 (0.86)& 23.10 (0.60) & 29.45 (1.03)\\
        
        \texttt{FedDecorr} (with \texttt{GC-Fed}) & 36.63 (3.14)& 44.08 (1.72)& 28.46 (1.28)& 35.04 (1.74)\\
        \midrule
        \texttt{FedSOL} & 19.15 (1.85)& 28.09 (0.87)& 15.03 (0.82) & 22.82 (0.23)\\
        
        \texttt{FedSOL} (with \texttt{Global GC}) & 27.36 (1.41) & 33.04 (1.35)& 20.69 (0.77)& 26.37 (0.79)\\
        
        \texttt{FedSOL} (\texttt{GC-Fed}) & 35.34 (2.09) & 42.01 (1.58)& 25.86 (1.24)& 31.69 (0.64) \\
        
        \bottomrule
    \end{tabular}
    }

    \label{tab:integration-experiment}
\end{table}

We assess the potential of our approach to enhance several baseline algorithms by integrating our GC-based methods. Focusing on proximal loss–based techniques, we select \texttt{FedProx}, \texttt{FedLC}, \texttt{FedNTD}, \texttt{FedDecorr}, and \texttt{FedSOL} to minimize conflicts when combining gradient projection with variance reduction strategies. As demonstrated in Table~\ref{tab:integration-experiment}, integrating \texttt{GC-Fed} leads to substantial improvements in performance across all methods. For example, in the 5/400, $\alpha=0.05$ setting, \texttt{FedProx} improves from 13.90\% to 36.30\% (a gain of 22.40 percentage points), while \texttt{FedLC} increases from 25.68\% to 39.91\% (a 14.23 percentage point improvement). Similar trends are observed in other settings, with GC-Fed augmentations yielding improvements typically ranging between 14 and 23 percentage points.

Furthermore, the incorporation of \texttt{Global GC}—which only requires server-side modifications—also enhances performance, albeit to a lesser extent. For instance, \texttt{FedDecorr} in the 5/400, $\alpha=0.05$ setting rises from 15.86\% to 26.55\% with \texttt{Global GC} (an increase of 10.69 percentage points), and comparable gains are evident in other configurations, often exceeding 10 percentage points.

Overall, our results indicate that GC-based algorithms, particularly \texttt{GC-Fed}, provide a robust performance boost, with average improvements in the range of 17–22 percentage points across various methods and settings. Notably, \texttt{FedLC} exhibits the highest absolute gain when integrated with \texttt{GC-Fed}, suggesting that our approach is especially beneficial for methods that already demonstrate strong baseline performance. These findings underscore the potential of our GC-based enhancements as a critical technique for advancing FL systems.

\subsubsection{CKA Similarity Analysis}
\begin{figure}[htbp] 
    \centering
    \includegraphics[width=\linewidth]{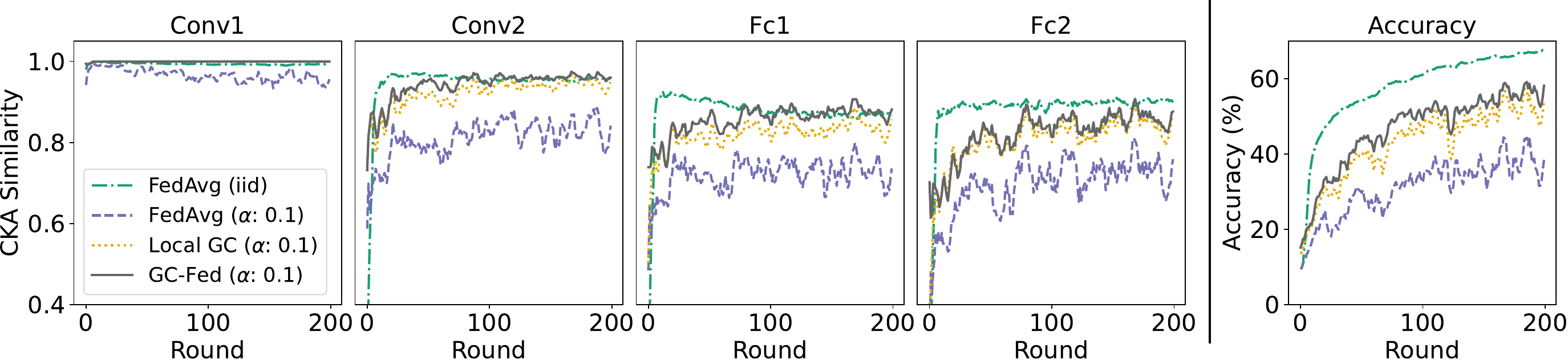}
\caption{Layer-wise averaged CKA similarity between representations of local models and the global model at each round (values range from 0 to 1, with higher values indicating greater similarity). Results are shown for CNN on CIFAR-10 with a participation ratio of 5/200.}

    \label{fig:CKA-sim}
\end{figure}

Beyond evaluating final performance, we also assess the similarity between client and global models using Centered Kernel Alignment (CKA) similarity~\cite{kornblith2019similarity}. CKA similarity is a metric that quantifies the alignment between representations by comparing the centered kernel matrices derived from their activations, thereby providing a scale-invariant measure of similarity across layers and widely adopted in FL research to compare local model representations~\cite{luo2021no,li2023effectiveness,kim2024communication}. In Figure~\ref{fig:CKA-sim}, we report the CKA similarity for each layer across four different scenarios. We treat \texttt{FedAvg} under the \textit{i.i.d.}  setting as the ideal reference because it yields higher layer-wise similarities compared to the \textit{non-i.i.d.} case. When \texttt{Local GC} is applied, the similarity increases, with the early layers nearly matching the ideal \texttt{FedAvg}-\textit{i.i.d.} similarity. Notably, when we employ \texttt{GC-Fed} (by applying \texttt{Global GC} to the \texttt{fc2} layer), the similarity improves not only in the \texttt{fc2} layer but also in the \texttt{conv2} and \texttt{fc1} layers, which correlates with an increase in accuracy.

\subsection{Ablation Studies}
\subsubsection{Choice of Mean Vector.}\label{sec:mu-vector} 
To evaluate the impact of the mean vector $\mu_{\mathbf{G}}$ (Equation~\eqref{eq:mu_g}) in GC, we specifically investigate whether dimensional choice of $\mu_{\mathbf{G}}$ is a crucial factor for performance improvements. To this end, we experiment with various methods for calculating the $\mu_{\mathbf{G}}$ vector. In GC, the $\mu$ vector computation spans either the output channels (for convolutional layers) or the output features (for fully connected layers). For a convolutional layer with shape $[C_{out}, C_{in}, K_{width}, K_{height}]$, the resulting $\mu$ vector has shape $[C_{out}, 1, 1, 1]$. Similarly, for a fully connected layer with shape $[F_{out}, F_{in}]$, the $\mu$ vector shape is $[F_{out}, 1]$.

\begin{figure}[htbp] 
    \centering
    \includegraphics[width=0.5\linewidth]{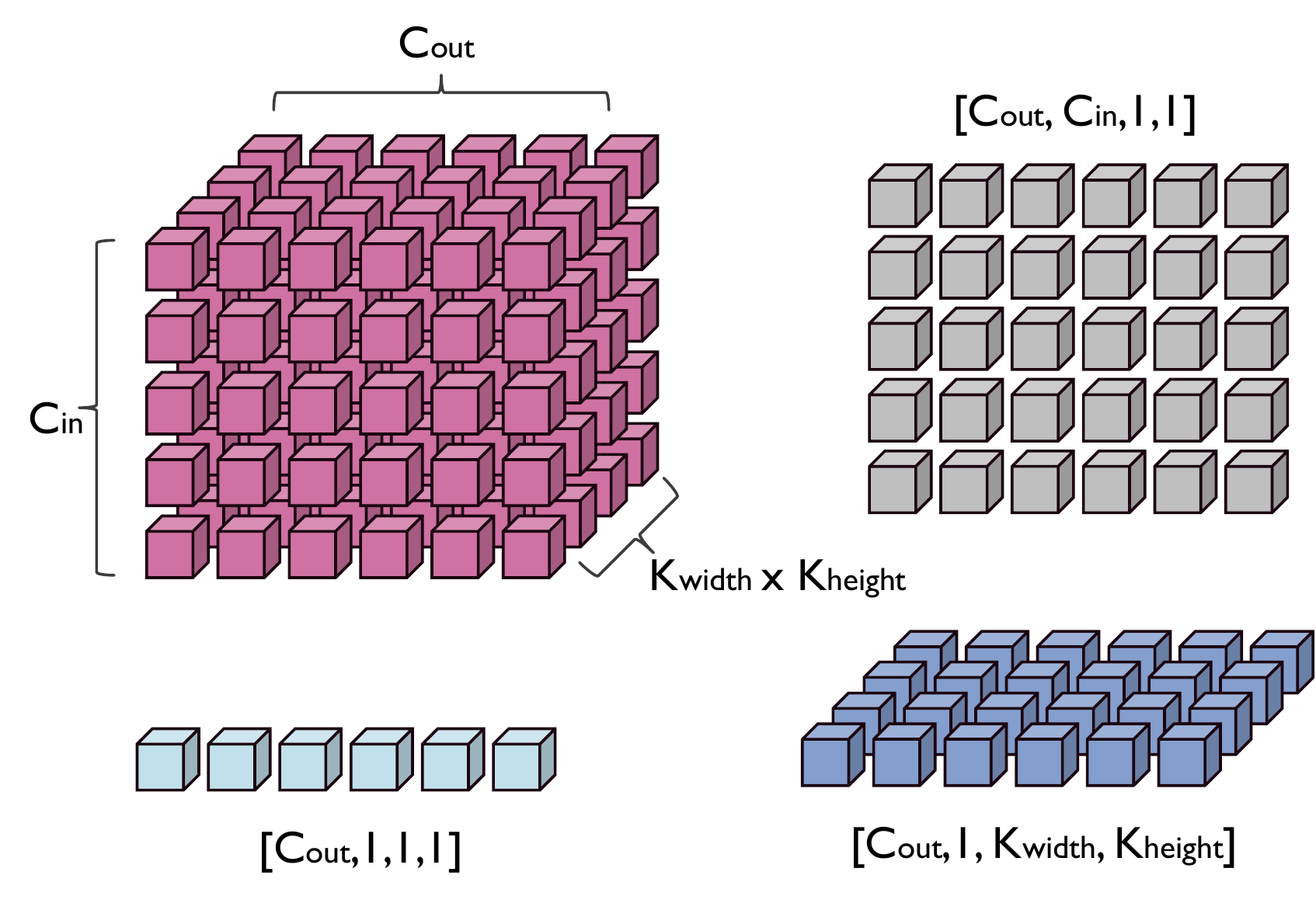}
    \caption{Illustration of various approaches for calculating the mean gradient matrix (vector) $\mu$ in a convolutional layer. To facilitate 3D visualization, the kernel width and height dimensions ($K_{width}$ $\times$ $K_{height}$) are combined.}
    \label{fig:mean-vector-slicing}
\end{figure}

\begin{table}[htbp]
\centering
\scriptsize
\caption{Top-1 accuracy comparison for different choices of the $\mu$ vector on CIFAR-10 with a CNN model. Experiments are conducted with a client participation ratio of 10/100 over 200 communication rounds, with results averaged over three independent runs. Bold text indicates the highest accuracy in each column, while underlined text represents the second highest accuracy.}

\resizebox{0.5\linewidth}{!}{
    \begin{tabular}{ccc}
        \toprule
        $\mu$ & LDA $\alpha$=0.1 & LDA $\alpha$=1.0 \\
        \toprule
        $[C_{out},1,1,1]$ & \textbf{61.50} ($\pm1.29$)& \textbf{70.51} ($\pm$0.11)\\
        
        $[C_{out},C_{in},1,1]$ & 58.33 ($\pm$1.49) & 68.57 ($\pm$0.40)\\
        
        $[C_{out},1,K_{w},K_{h}]$ & 59.28 ($\pm$1.39)& 67.00 ($\pm$0.42)\\
        
        $[C_{out},C_{in},1,K_{h}]$ & 56.20 ($\pm$0.91)& 66.43 ($\pm$0.36)\\
        
        $[1,C_{in},K_{w},K_{h}]$ & \underline{60.31} ($\pm$1.90)&\underline{69.87} ($\pm$0.49) \\
        \midrule
        \texttt{FedAvg} & 52.22 ($\pm$0.90)&62.50 ($\pm$1.24) \\
        \bottomrule
    \end{tabular}
    }
    \label{tab:mean-vector-performance}
\end{table}
We have conducted an ablation study to evaluate the impact of various $\mu$ configurations, as depicted in Figure~\ref{fig:mean-vector-slicing}. Notably, these adjustments are restricted to the convolutional layer. The results, summarized in Table~\ref{tab:mean-vector-performance}, indicate that while the original GC-based method of calculating \(\mu\) achieves the best performance, it also exhibits minimal performance variation across different \(\mu\) values, consistently surpassing \texttt{FedAvg}. Furthermore, even when the output dimension $[1,C_{in},K_{w},K_{h}]$ is excluded, performance remains robust. This result verifies that the regularization with the mean vector $\mu$ has consistent performance enhancement while the output-channel-wise mean vector shows the highest improvements.

\subsubsection{Layer Borderline $\lambda$} \label{sec:layer-borderline}
\begin{figure*}[htbp]
    \centering
    \includegraphics[width=0.95\linewidth]{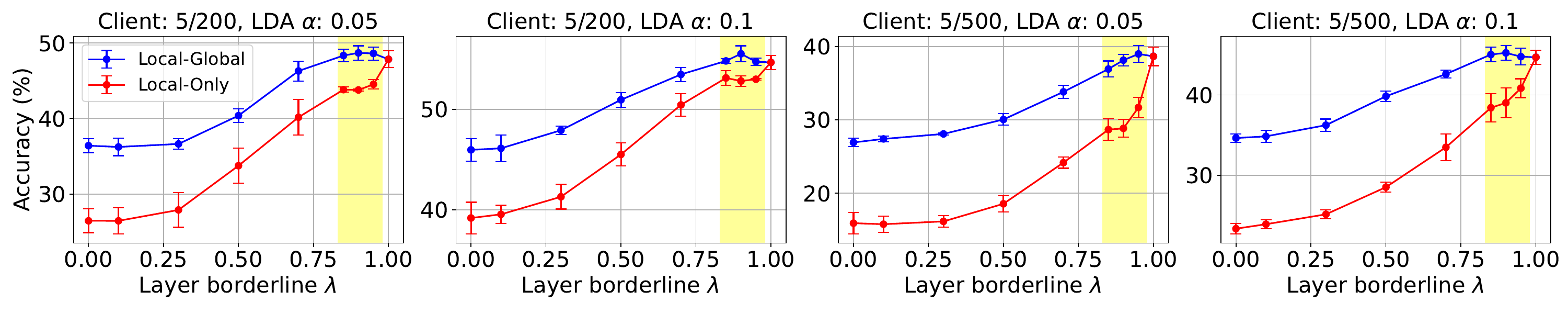}
\caption{Top-1 test accuracy comparison for different ratios ($\lambda$) of \texttt{Local GC} and \texttt{Global GC}, as well as Local GC only. Results are reported for ResNet18 on CIFAR-100 over 1000 rounds.}

    \label{fig:ablation-lambda}
\end{figure*}
To assess whether \texttt{GC-Fed} is applicable more flexibly, we conduct experiments exploring various borderline settings by decoupling \texttt{Local GC} from \texttt{Global GC}. Specifically, a $\lambda$ value of 0 denotes the application of \texttt{Global GC}, while a $\lambda$ value of 1 corresponds to \texttt{Local GC}. We further compare performance when layers beyond the designated borderline are processed with \texttt{Global GC} against the case where only layers up to the borderline receive \texttt{Local GC}. Note that although ResNet18 nominally comprises 18 layers, its implementation—separating weights, biases, and including a group normalization layer—yields 62 learnable parameter groups.

Figure~\ref{fig:ablation-lambda} shows that shifting the borderline further into the latter layers consistently improves performance. Notably, our experiments across varying client counts and \textit{non-i.i.d.} intensities demonstrate that configurations with $\lambda$ values in the range [0.85, 0.95] outperform the full \texttt{Local GC} configuration. This finding suggests that determining an appropriate borderline within the latter layers and applying \texttt{Global GC} to these layers enables \texttt{GC-Fed} to adapt more effectively to diverse FL environments and DNN architectures.

Conversely, when only \texttt{Local GC} is employed (i.e., with $\lambda=0$ representing \texttt{FedAvg} and $\lambda=1$ representing \texttt{Local GC}), performance remains consistently inferior. In summary, while applying GC across all layers is beneficial, supplementing the latter layers with \texttt{Global GC} yields.

\subsubsection{Impact of Hyperparameters}
\begin{figure*}[htbp]
    \centering
    \includegraphics[width=\linewidth]{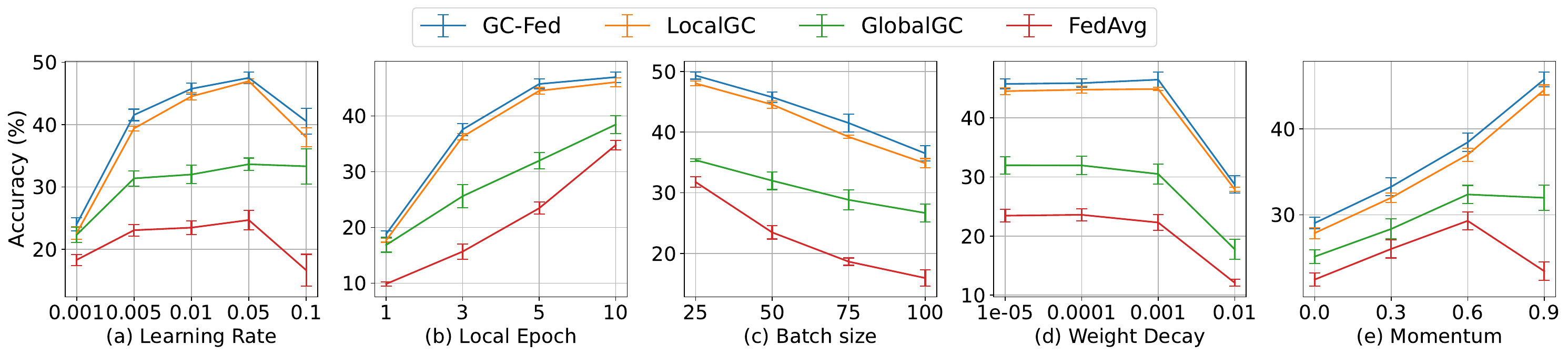}
\caption{Top-1 accuracy of the GC-based algorithm and \texttt{FedAvg} under varying hyperparameter settings. Results are reported for ResNet18 on CIFAR-100 with 400 communication rounds, a participation ratio of 5/200, and LDA $\alpha = 0.1$. Each experiment is averaged over three independent runs, with error bars representing the standard deviation.}

    \label{fig:ablation}
\end{figure*}
In Figure~\ref{fig:ablation}, we present an ablation study to examine the effect of different hyperparameters on the performance of \texttt{GC-Fed}, in comparison to \texttt{FedAvg} and other GC-based methods. Unless otherwise stated, the default hyperparameter values are: learning rate of 0.01, local epochs set to 5, batch size of 50, weight decay of $1\times10^{-5}$, and momentum coefficient of 0.9. Each plot in the figure modifies a single hyperparameter while keeping the others constant.

For the learning rate, performance improves across all algorithms as it increases up to 0.05. However, when set to 0.1, the performance drops significantly with higher variance within a given round, indicating an instability in training. Regarding the number of local epochs and batch size, we observe that increasing the number of local epochs while reducing the batch size leads to better performance. This is likely because a lower batch size and more local updates result in a higher number of overall parameter updates per client, leading to more refined local models~\cite{mcmahan2017communication,seo2024understanding}.

Weight decay has minimal impact on performance unless it is excessively high, in which case it disrupts training and degrades performance. The most interesting trend is observed with the momentum coefficient. Both \texttt{Local GC} and \texttt{GC-Fed} benefit from a higher momentum coefficient, leading to better performance. However, for \texttt{Global GC} and \texttt{FedAvg}, a high momentum coefficient (0.9) negatively affects performance. This discrepancy suggests that applying GC locally—especially in the early layers—enables the training procedure to more effectively exploit momentum SGD even in \textit{non-i.i.d.} settings.

Nevertheless, across all hyperparameter settings, \texttt{GC-Fed} consistently outperforms \texttt{Local GC}, demonstrating its effectiveness. This suggests that leveraging deeper layers (e.g., classifier layers) with \texttt{Global GC} contributes to additional performance gains when applying GC in FL.

\section{Conclusion}

In this work, we investigate methods to enhance FL performance in challenging scenarios characterized by high data heterogeneity and severe partial client participation. Our focus centers on techniques for gradient adjustment. Many existing approaches improve performance by modifying the local gradient update process using reference information derived from previous rounds—information that must be either stored or transmitted. Moreover, these methods often struggle under severe partial participation due to the variability in client selection across rounds. To address this limitation, we introduce a reference-free gradient correction FL framework \texttt{GC-Fed} through GC, applying it selectively on a layer-by-layer basis in both local and global stages. This approach enables us to achieve robust and high performance, even in the highly adverse conditions commonly encountered in FL environments.

\appendix

\section{Theoretical Analysis}

\textbf{Proof sketch} of Lemma \ref{lemma1}

Here, we show how GC can approach faster than \texttt{FedAvg} algorithm to the optimum point.
We compare one-step gradient of \texttt{FedAvg} and \texttt{GC-Fed} and show that \texttt{GC-Fed} further reduces the gap to the optimum point $\mathbf{w}^*=\arg \min F(\mathbf{w})$ than \texttt{FedAvg}.
$L_2$ distance of $\mathbf{w}^*$ and update of the weight $\mathbf{w}^t$ is compared for \texttt{FedAvg} and \texttt{GC-Fed} and will be shown than distance becomes smaller with \texttt{GC-Fed} update than with \texttt{FedAvg}.

\subsection{Proof of Lemma 1}
\begin{proof}
The update of local weight at update step $\tau$ can be expressed as follows.

\begin{equation}\label{update}
    \mathbf{w}_k^{\tau+1} = \mathbf{w}_k^\tau - \eta_\tau \nabla F_k(\mathbf{w}_k^\tau , \xi_k^\tau) 
\end{equation}
Global synchronization step is $\mathcal{I}_E=\{ nE|n=1,2, \cdots \}$, the weights are uploaded to server only when $\tau+1 \in \mathcal{I}_E$. In other words, every $E$ local updates equals to a round and we have a single model aggregation per round.
For convenience, we let $\overline{\mathbf{w}}^\tau=\sum_{k=1}^N p_k \mathbf{w}_k^\tau $, $\mathbf{G}_\tau = \sum_{k=1}^N p_k \nabla F_k (\mathbf{w}_k^\tau, \xi_k^\tau)$ and
$\overline{\mathbf{G}}_\tau=\sum_{k=1}^N p_k \nabla F_k(\mathbf{w}_k^\tau) $, thus $\mathbb{E}[\mathbf{G}_\tau]=\overline{\mathbf{G}}_\tau$.

In the gradient projection-based FL, the projected gradient $\Tilde{\mathbf{G}}^t$ is used instead of the gradient $\mathbf{G}^t$. 
\begin{equation}
    \mathbf{P}\mathbf{G}^t=\Tilde{\mathbf{G}}^t 
\end{equation}
\begin{equation}
    \mathbf{P}\overline{\mathbf{G}}^t=\overline{\Tilde{\mathbf{G}}}^t
\end{equation}
Then, the update rule of local weight (Eq. \ref{update}) for the \texttt{GC-Fed} can be expressed as follows.

\begin{equation}
\mathbf{w}_k^{\tau+1} = \mathbf{w}_k^\tau - \eta_\tau \mathbf{P} \mathbf{G}_k^\tau = \mathbf{w}_k^\tau - \eta_\tau \Tilde{\mathbf{G}}^\tau
\end{equation}
Note that this can also be used as update rule for \textbf{Global} \texttt{GC-Fed} in case of one local update per round scenario.
The result of one step SGD for \texttt{FedAvg} and \texttt{GC-Fed} can be summarized as follows.

\begin{equation}
    \texttt{FedAvg}:\mathbf{w}_k^{\tau+1} = \mathbf{w}_k^\tau - \eta_\tau \mathbf{G}^\tau , ~~~~\texttt{GC-Fed}:\mathbf{w}_k^{\tau+1} = \mathbf{w}_k^\tau - \eta_\tau \Tilde{\mathbf{G}}^\tau 
\end{equation}

Our goal is to show that the \textit{gap} of $\tau+1$ step weight $\overline{\mathbf{w}}_{\tau+1}$ and the optimum weight $\mathbf{w}^*$ becomes smaller in \texttt{GC-Fed} than \texttt{FedAvg}. We denote \textit{gap} as the squared $L_2$ distance.
Using the update rule, squared $L_2$ distance between $\overline{\mathbf{w}}_{\tau+1}$ and the optimum weight $\mathbf{w}^*$ is
\begin{equation}\label{gap}
\begin{aligned}
\| \overline{\mathbf{w}}_{\tau+1} - \mathbf{w}^* \|^2 &= \| \overline{\mathbf{w}}_\tau - \eta_\tau \Tilde{\mathbf{G}}_\tau  - \mathbf{w}^* \|^2 \\
&= \| \overline{\mathbf{w}}_\tau - \mathbf{w}^* - \eta_\tau \Tilde{\mathbf{G}}_\tau  - \eta_\tau \overline{\Tilde{\mathbf{G}}}_\tau + \eta_\tau \overline{\Tilde{\mathbf{G}}}_\tau \|^2 \\
&= \underbrace{\| \overline{\mathbf{w}}_\tau - \mathbf{w}^* - \eta_\tau \overline{\Tilde{\mathbf{G}}}_\tau  \|^2}_{A_1} + \underbrace{\eta_\tau^2 \| \Tilde{\mathbf{G}}_\tau - \overline{\Tilde{\mathbf{G}}}_\tau \|^2 }_{A_2}
\\
&\quad + \underbrace{2\eta_\tau \langle \overline{\mathbf{w}}_\tau - \mathbf{w}^* - \eta_\tau \overline{\Tilde{\mathbf{G}}}_\tau , \Tilde{{\mathbf{G}}}_\tau - \overline{\Tilde{\mathbf{G}}}_\tau \rangle }_{A_3}.
\end{aligned}
\end{equation}
For the case of \texttt{FedAvg}, we can just replace the gradient $\Tilde{\mathbf{G}}$ with $\mathbf{G}$.
The first term $A_1$ can be rearranged with the gap of previous step and the remainder as follows.

\begin{equation}
\begin{aligned}
    &A_1: \| \overline{\mathbf{w}}_\tau - \mathbf{w}^* - \eta_\tau \overline{\Tilde{\mathbf{G}}}_\tau  \|^2 \\
    &=\| \overline{\mathbf{w}}_\tau - \mathbf{w}^*\|^2 
    \underbrace{-2\eta_\tau \langle\overline{\mathbf{w}}_\tau - \mathbf{w}^*, \overline{\Tilde{\mathbf{G}}}_\tau \rangle}_{B_1} + \underbrace{\eta_\tau^2\|\overline{\Tilde{\mathbf{G}}}_\tau\|^2}_{B_2}     
\end{aligned}    
\end{equation}

Note that $\mathbb{E}[A_3]=0$ due to unbiased gradient.
Therefore, writing the terms altogether, the \textit{gap} between $\overline{\mathbf{w}}_{\tau+1}$ and the optimum weight $\mathbf{w}^*$ can be expressed as follows. 
\begin{equation}
\begin{aligned}
    \| \overline{\mathbf{w}}_{\tau+1} - \mathbf{w}^* \|^2 = \| \overline{\mathbf{w}}_\tau - \mathbf{w}^*\|^2 
    \underbrace{-2\eta_\tau \langle\overline{\mathbf{w}}_\tau - \mathbf{w}^*, \overline{\Tilde{\mathbf{G}}}_\tau \rangle}_{B_1} + \underbrace{\eta_\tau^2\|\overline{\Tilde{\mathbf{G}}}_\tau\|^2}_{B_2}+\underbrace{\eta_\tau^2 \| \Tilde{\mathbf{G}}_\tau - \overline{\Tilde{\mathbf{G}}}_\tau \|^2 }_{A_2}
\end{aligned}    
\end{equation}
Therefore, we can see that one step update of Projected gradient reduces the gap between $\overline{\mathbf{w}}_{\tau+1}$ and the optimum weight $\mathbf{w}^*$ compared with the gap between $\overline{\mathbf{w}}_{\tau}$ and the optimum weight $\mathbf{w}^*$ by $\eta_\tau^2\|\overline{\Tilde{\mathbf{G}}}_\tau\|^2 + \eta_\tau^2 \| \Tilde{\mathbf{G}}_\tau - \overline{\Tilde{\mathbf{G}}}_\tau \|^2 -2\eta_\tau \langle\overline{\mathbf{w}}_\tau - \mathbf{w}^*, \overline{\Tilde{\mathbf{G}}}_\tau \rangle$.

\end{proof}

\textbf{Proof sketch} of Theorem \ref{th1}

To show the faster gap reduction than \texttt{FedAvg} algorithm, we compare the gap of \texttt{FedAvg} and \texttt{GC-Fed} by corresponding terms. For instance, \texttt{FedAvg} has corresponding term of $A_1$ in Eq. \ref{gap} if the gradient $\Tilde{\mathbf{G}}$ is replaced with $\mathbf{G}$.

\subsection{Proof of Theorem 1}
\begin{proof}
\hfill

1) Comparison of $A_1$ 

% A1 comparison
\begin{align*}
  \begin{split}
  &\texttt{FedAvg}\\
  &A_1: \| \overline{\mathbf{w}}_\tau - \mathbf{w}^* - \eta_\tau \overline{\mathbf{G}}_\tau  \|^2 \\ 
  &=\| \overline{\mathbf{w}}_\tau - \mathbf{w}^*\|^2 
  \underbrace{-2\eta_\tau \langle\overline{\mathbf{w}}_\tau - \mathbf{w}^*, \overline{\mathbf{G}}_\tau \rangle}_{B_1} + 
  \underbrace{\eta_\tau^2\|\overline{\mathbf{G}}_\tau\|^2}_{B_2}
  \end{split}
\quad\quad
  \begin{split}
  &\texttt{GC-Fed}\\
  &A_1: \| \overline{\mathbf{w}}_\tau - \mathbf{w}^* - \eta_\tau \overline{\Tilde{\mathbf{G}}}_\tau  \|^2 \\
  &=\| \overline{\mathbf{w}}_\tau - \mathbf{w}^*\|^2 
  \underbrace{-2\eta_\tau \langle\overline{\mathbf{w}}_\tau - \mathbf{w}^*, \overline{\Tilde{\mathbf{G}}}_\tau \rangle}_{B_1} + \underbrace{\eta_\tau^2\|\overline{\Tilde{\mathbf{G}}}_\tau\|^2}_{B_2} 
  \end{split}
\end{align*}

The first terms of $A_1$ in \texttt{FedAvg} and \texttt{GC-Fed} are equivalent.
We can compare $B_1$ by subtracting $B_1$ of \texttt{GC-Fed} from \texttt{FedAvg}.

\begin{align*}
    B_{1, \textrm{FA}}-B_{1, \textrm{GC}}=&-2\eta_\tau \langle\overline{\mathbf{w}}_\tau - \mathbf{w}^*, \overline{\mathbf{G}}_\tau \rangle - (-2\eta_\tau \langle\overline{\mathbf{w}}_\tau - \mathbf{w}^*, \mathbf{P}\overline{\mathbf{G}}_\tau \rangle ) \\
    &= -2\eta_\tau \langle\overline{\mathbf{w}}_\tau - \mathbf{w}^*, \overline{\mathbf{G}}_\tau \rangle +2\eta_\tau \langle\overline{\mathbf{w}}_\tau - \mathbf{w}^*, \overline{\mathbf{G}}_\tau \rangle
    -2\eta_\tau \langle\overline{\mathbf{w}}_\tau - \mathbf{w}^*, \mathbf{e}\mathbf{e}^\intercal \overline{\mathbf{G}} \rangle \\
    &= -2\eta_\tau \langle\overline{\mathbf{w}}_\tau - \mathbf{w}^*, \mathbf{e}\mathbf{e}^\intercal \overline{\mathbf{G}}_\tau\rangle
\end{align*}

If the assumption \ref{AS_ortho} holds, $\mathbf{e}^\intercal \mathbf{w}^* = 0$. Also, since the weights are always projected to orthogonal subspace, the previous weight satisfies $\mathbf{e}^\intercal \overline{\mathbf{w}}_\tau = 0$, therefore,  $B_{1, \textrm{FA}}-B_{1, \textrm{PG}}=0$. 

% B2 comparison
\begin{align*}
  \begin{split}
  &\texttt{FedAvg}\\
  &B_2: \eta_\tau^2\|\overline{\mathbf{G}}_\tau\|^2 
  \end{split}
\quad\quad
  \begin{split}
  &\texttt{GC-Fed}\\
  &B_2: \eta_\tau^2\|\overline{\Tilde{\mathbf{G}}}_\tau\|^2 \\
  &=\eta_\tau^2 (\|\overline{\mathbf{G}}_\tau\|^2 -\|\mathbf{e}^\intercal \overline{\mathbf{G}}_\tau \|^2)
  \end{split}
\end{align*}

$B_2$ of \texttt{GC-Fed} is smaller than \texttt{FedAvg} by $\eta_\tau^2\|\mathbf{e}^\intercal \mathbf{G}_\tau \|^2$.
Therefore, combining $B_1$ and $B_2$, we can see $A_{1, \textrm{GC}}$ is smaller than $A_{1, \textrm{FA}}$ by nonnegative $\eta_\tau^2\|\mathbf{e}^\intercal \mathbf{G}_\tau \|^2$.

2) Comparison of $A_2$

% A2 comparison
\begin{align*}
  \begin{split}
  &\texttt{FedAvg}\\
  &A_2: \eta_\tau^2 \| \mathbf{G}_\tau - \overline{\mathbf{G}}_\tau \|^2 \\ 
  & ~~\\
  & ~~
  \end{split}
\quad\quad
  \begin{split}
  &\texttt{GC-Fed}\\
  &A_2: \eta_\tau^2 \| \Tilde{\mathbf{G}}_\tau - \overline{\Tilde{\mathbf{G}}}_\tau \|^2 \\
  &= \eta_\tau^2 \| \mathbf{P}\mathbf{G}_\tau - \mathbf{P}\overline{\mathbf{G}}_\tau \|^2\\
  &= \eta_\tau^2 \| \mathbf{G}_\tau - \overline{\mathbf{G}}_\tau \|^2 - \eta_\tau^2 \|\mathbf{e}^\intercal( \mathbf{G}_\tau - \overline{\mathbf{G}}_\tau) \|^2
  \end{split}
\end{align*}

As can be seen, $A_2$ of \texttt{GC-Fed} is smaller than \texttt{FedAvg} by nonnegative $\eta_\tau^2 \|\mathbf{e}^\intercal( \mathbf{G}_\tau - \overline{\mathbf{G}}_\tau) \|^2$.

3) $\mathbb{E}[A_3]=0$ due to unbiased gradient which results in $A_3$ to be eliminated for the both algorithms.

\end{proof}

\subsection{Bounding the discrepancy}
Here, we describe the potential discrepancy in the case where the assumption \ref{AS_ortho} does not hold. 
% In practice, we normally observe very small discrepancy between $F(\mathbf{w}^*) - F(\mathbf{w}^*_{\parallel})$.
As stated in Eq. \ref{po}, we can decompose $\mathbf{w}^*$ into parallel and orthogonal components as follows.
\begin{equation}
\mathbf{w}^* = \mathbf{w}^*_{\parallel} + \mathbf{w}^*_{\perp}
\end{equation}
where $\mathbf{w}^*_{\parallel} = \mathbf{P} \mathbf{w}^*$ is the component in the subspace and $\mathbf{w}^*_{\perp} = (\mathbf{I} - \mathbf{P}) \mathbf{w}^*$ is the component orthogonal to subspace.
The difference between the true optimum and the convergence point is
\begin{equation}
    \Delta = \mathbf{w}^* - \mathbf{w}^*_{\parallel} = \mathbf{w}^*_{\perp} = \mathbf{e}\mathbf{e}^\intercal \mathbf{w}^*.
\end{equation}
This is simply a column-wise mean vector of $\mathbf{w}$ which converges to $0$ with the law of large numbers with random unbiased weights.
With following assumption on the L-smoothness of the loss function we can bound the discrepancy.
\begin{assumption}\label{AS1}
\textit{L-Smoothness: For all $\mathbf{w}$ and $\mathbf{w'}$, $F_k(\mathbf{w})\leq F_k(\mathbf{w'})+(\mathbf{w}-\mathbf{w'})^\intercal\nabla F_k(\mathbf{w'})+\frac{L}{2} || \mathbf{w}-\mathbf{w'}||_2^2$}
\end{assumption}

The residual error due to the discrepancy is $F(\mathbf{w}^*) - F(\mathbf{w}^*_{\parallel})$ which is bounded by $\| \mathbf{w}^*_{\parallel} - \mathbf{w}^* \|^2 = \| \Delta \|$ with L-smoothness condition as follows.
\begin{equation}
    F(\mathbf{w}^*_{\parallel}) - F(\mathbf{w}^*) \leq \frac{L}{2} \| \Delta \|^2.
\end{equation}

\printcredits

% \bibliographystyle{cas-model2-names}
\bibliographystyle{unsrt}

\bibliography{main}

\end{document}